# Neon2: Finding Local Minima via First-Order Oracles
(version 3)

Zeyuan Allen-Zhu
zeyuan@csail.mit.edu
Microsoft Research AI

Yuanzhi Li
yuanzhil@cs.princeton.edu
Princeton University

November 17, 2017[*]

## Abstract

We propose a reduction for non-convex optimization that can (1) turn an stationary-point finding algorithm into an local-minimum finding one, and (2) replace the Hessian-vector product computations with only gradient computations. It works both in the stochastic and the deterministic settings, without hurting the algorithm's performance.

As applications, our reduction turns Natasha2 into a first-order method without hurting its performance. It also converts SGD, GD, SCSG, and SVRG into algorithms finding approximate local minima, outperforming some best known results.

## 1 Introduction

Nonconvex optimization has become increasingly popular due its ability to capture modern machine learning tasks in large scale. Most notably, training deep neural networks corresponds to minimizing a function

$$f(x) = \frac{1}{n}\sum_{i=1}^{n} f_i(x)$$

over $x \in \mathbb{R}^d$ that is non-convex, where each training sample $i$ corresponds to one loss function $f_i(\cdot)$ in the summation. This average structure allows one to perform stochastic gradient descent (SGD) which uses a random $\nabla f_i(x)$ —corresponding to computing backpropagation once— to approximate $\nabla f(x)$ and performs descent updates.

Motivated by such large-scale machine learning applications, we wish to design faster first-order non-convex optimization methods that outperform the performance of gradient descent, both in the *online* and *offline* settings. In this paper, we say an algorithm is online if its complexity is independent of $n$ (so $n$ can be infinite), and offline otherwise. In recently years, researchers across different communities have gathered together to tackle this challenging question. By far, known theoretical approaches mostly fall into one of the following two categories.

**First-order methods for stationary points.** In analyzing first-order methods, we denote by gradient complexity $T$ the number of computations of $\nabla f_i(x)$. To achieve an $\varepsilon$-approximate stationary point —namely, a point $x$ with $\|\nabla f(x)\| \leq \varepsilon$— it is a folklore that gradient descent (GD) is offline and needs $T \propto O\left(\frac{n}{\varepsilon^2}\right)$, while stochastic gradient decent (SGD) is online and needs $T \propto O\left(\frac{1}{\varepsilon^4}\right)$.

---

[*]V1 appeared on this date and V2/V3 improves writing. While we were preparing V1 of this paper, we were informed the similar work of Xu and Yang [31]. To respect the fact that their work appeared online slightly before us, we have adopted their algorithm name Neon and called our new algorithm Neon2. We encourage readers citing this work to also cite [31].

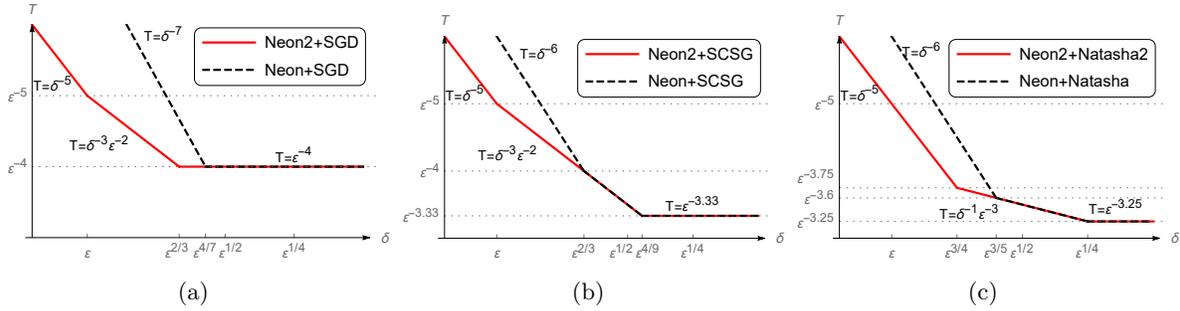

Figure 1: `Neon` vs `Neon2` for finding $(\varepsilon, \delta)$-approximate local minima. We emphasize that `Neon2` and `Neon` are based on the same high-level idea, but `Neon` is arguably the first-recorded result to turn stationary-point finding algorithms (such as `SGD`, `SCSG`) into local-minimum finding ones, with theoretical proofs.

In recent years, the offline complexity has been improved to $T \propto O\left(\frac{n^{2/3}}{\varepsilon^2}\right)$ by the `SVRG` method [4, 24], and the online complexity has been improved to $T \propto O\left(\frac{1}{\varepsilon^{10/3}}\right)$ by the `SCSG` method [19]. Both of them rely on the so-called variance-reduction technique, originally discovered for convex problems [12, 17, 27, 29].

Both algorithms `SVRG` and `SCSG` are only capable of finding approximate *stationary points*, which may not necessarily be approximate local minima and are arguably bad solutions for deep neural nets [10, 11, 15]. Thus,

> *can we turn stationary-point finding algorithms into local-minimum finding ones?*

**Hessian-vector methods for local minima.** It is common knowledge that using information about the Hessian, one can find $\varepsilon$-approximate local minima —namely, a point $x$ with $\|\nabla f(x)\| \leq \varepsilon$ and also $\nabla^2 f(x) \succeq -\varepsilon^{1/C} \mathbf{I}$.[1] In 2006, Nesterov and Polyak [21] showed that one can find an $\varepsilon$-approximate in $O(\frac{1}{\varepsilon^{1.5}})$ iterations, but each iteration requires an (offline) computation as heavy as inverting the matrix $\nabla^2 f(x)$.

To fix this issue, researchers propose to study the so-called "Hessian-free" methods that, in addition to gradient computations, also compute Hessian-vector products. That is, instead of using the full matrix $\nabla^2 f_i(x)$ or $\nabla^2 f(x)$, these methods also compute $\nabla^2 f_i(x) \cdot v$ for indices $i$ and vectors $v$.[2] For Hessian-free methods, we denote by gradient complexity $T$ the number of computations of $\nabla f_i(x)$ plus that of $\nabla^2 f_i(x) \cdot v$. The hope of using Hessian-vector products is to improve the complexity $T$ as a function of $\varepsilon$.

Such improvement was first shown possible independently by [1, 8] for the offline setting, with complexity $T \propto \left(\frac{n}{\varepsilon^{1.5}} + \frac{n^{3/4}}{\varepsilon^{1.75}}\right)$ so is better than that of gradient descent. In the online setting, the first improvement was by `Natasha2` which gives complexity $T \propto \left(\frac{1}{\varepsilon^{3.25}}\right)$ [2].

Unfortunately, it is argued by some researchers that Hessian-vector products are not general enough and may not be as simple to implement as evaluating gradients [9]. Therefore,

> *can we turn Hessian-free methods into first-order ones, without hurting their performance?*

---
[1] We say $\mathbf{A} \succeq -\delta \mathbf{I}$ if all the eigenvalues of $\mathbf{A}$ are no smaller than $-\delta$. In this high-level introduction, we focus only on the case when $\delta = \varepsilon^{1/C}$ for some constant $C$.

[2] Hessian-free methods are useful because when $f_i(\cdot)$ is explicitly given, computing its gradient is in the same complexity as computing its Hessian-vector product [23] [28], using backpropagation.

## 1.1 From Hessian-Vector Products to First-Order Methods

Recall by definition of derivative we have $\nabla^2 f_i(x) \cdot v = \lim_{q \to 0} \{\frac{\nabla f_i(x+qv) - \nabla f_i(x)}{q}\}$. Given any Hessian-free method, at least at a high level, can we replace every occurrence of $\nabla^2 f_i(x) \cdot v$ with $w = \frac{\nabla f_i(x+qv) - \nabla f_i(x)}{q}$ for some small $q > 0$?

Note the error introduced in this approximation is $\|\nabla^2 f_i(x) \cdot v - w\| \propto q\|v\|^2$. However, the original algorithm might not be stable to adversarial noise, thus, an (inverse) exponentially small $q$ might be required. One of our main contributions is to show how to implement these algorithms stably, so we can convert Hessian-free methods into first-order ones with an (inverse) *polynomially small* $q$.

In this paper, we demonstrate this idea by converting *negative-curvature-search (NC-search)* subroutines into first-order processes. NC-search is a key subroutine used in state-of-the-art Hessian-free methods that have rigorous proofs [1, 2, 8]. It solves the following simple task:

---
**negative-curvature search (NC-search)**

given point $x_0$, decide if $\nabla^2 f(x_0) \succeq -\delta \mathbf{I}$ or find a unit vector $v$ such that $v^\top \nabla^2 f(x_0) v \leq -\frac{\delta}{2}$.

---

### 1.1.1 Online Setting

In the online setting, NC-search can be solved by Oja's algorithm [22] which costs $\widetilde{O}(1/\delta^2)$ computations of Hessian-vector products. This is first proved by Allen-Zhu and Li [7] and first applied to NC-search in Natasha2 [2]).

In this paper, we propose a method Neon2$^{\text{online}}$ which solves the NC-search problem via only stochastic first-order updates. That is, starting from $x_1 = x_0 + \xi$ where $\xi$ is some random perturbation, we keep updating $x_{t+1} = x_t - \eta(\nabla f_i(x_t) - \nabla f_i(x_0))$. In the end, the vector $x_T - x_0$ gives us enough information about the negative curvature.

**Theorem 1** (informal). *Our Neon2$^{\text{online}}$ algorithm solves NC-search using $\widetilde{O}(1/\delta^2)$ stochastic gradients, without Hessian-vector product computations.*

This complexity $\widetilde{O}(1/\delta^2)$ matches that of Oja's algorithm, and is information-theoretically optimal (up to log factors), see the lower bound in [7].

We emphasize that the independent work Neon by Xu and Yang [31] is actually the *first* recorded theoretical result that proposed this approach. However, Neon needs $\widetilde{O}(1/\delta^3)$ stochastic gradients, because it uses full gradient descent to find NC (on a sub-sampled objective) inspired by power method and [16]; instead, Neon2$^{\text{online}}$ uses *stochastic gradients* and is based on the recent result of Oja's algorithm [7].

Plugging Neon2$^{\text{online}}$ into Natasha2 [2], we achieve the following corollary (see Figure 1(c)):

**Theorem 2** (informal). *Neon2$^{\text{online}}$ turns Natasha2 into a stochastic first-order method, without hurting its performance. That is, it finds an $(\varepsilon, \delta)$-approximate local minimum in $T = \widetilde{O}(\frac{1}{\varepsilon^{3.25}} + \frac{1}{\varepsilon^3 \delta} + \frac{1}{\delta^5})$ stochastic gradient computations, without Hessian-vector product computations.*

(We say $x$ is an $(\varepsilon, \delta)$-approximate local minimum if $\|\nabla f(x)\| \leq \varepsilon$ and $\nabla^2 f(x) \succeq -\delta \mathbf{I}$.)

### 1.1.2 Offline Deterministic Setting

There are a number of ways to solve the NC-search problem in the offline setting using Hessian-vector products. Most notably, power method uses $\widetilde{O}(n/\delta)$ computations of Hessian-vector products, and Lanscoz method [18] uses $\widetilde{O}(n/\sqrt{\delta})$ computations.

In this paper, we convert (a variant of) Lanscoz's method into a first-order one:

**Theorem 3** (informal). *Our* `Neon2`[det] *algorithm solves NC-search using* $\widetilde{O}(1/\sqrt{\delta})$ *full gradients (or equivalently* $\widetilde{O}(n/\sqrt{\delta})$ *stochastic gradients).*

Although analyzed in the online setting only, the work `Neon` by Xu and Yang [31] also applies to the offline setting, and seems to be the *first* result to solve NC-search using first-order gradients with a theoretical proof. However, `Neon` needs $\widetilde{O}(1/\delta)$ full gradients. Their approach is inspired by [16], but our `Neon2`[det] is based on the stable Chebyshev approximation theory [6].

By putting `Neon2`[det] and `Neon2`[finite] into the `CDHS` method of Carmon et al. [8], we have

**Theorem 4** (informal). `Neon2`[det] *turns* `CDHS` *into a first-order method without hurting its performance: it finds an* $(\varepsilon, \delta)$-*approximate local minimum in* $\widetilde{O}\big(\frac{1}{\varepsilon^{1.75}} + \frac{1}{\delta^{3.5}}\big)$ *full gradient computations.*

One should perhaps compare `Neon2`[det] to the "convex until guilty" method by Carmon et al. [9]. Their method finds $\varepsilon$-approximate stationary points using $\widetilde{O}(1/\varepsilon^{1.75})$ full gradients, and is perhaps the first *first-order* method achieving a convergence rate better than $1/\varepsilon^2$ of `GD`. Their method may not guarantee approximate local minima. In comparison, `Neon2`[det] on `CDHS` achieves the same complexity but guarantees finding approximate local minima.

### 1.1.3 Offline Finite-Sum Setting

Recall one can also solve the NC-search problem in the offline setting by the (finite-sum) shift-and-invert [13] method, using $\widetilde{O}(n + n^{3/4}/\sqrt{\delta})$ computations of Hessian-vector products. We refer to this method as "finite-sum SI", and also convert it into a first-order method.

**Theorem 5** (informal). `Neon2`[finite] *algorithm solves NC-search using* $\widetilde{O}(n + n^{3/4}/\sqrt{\delta})$ *stochastic gradients.*

Putting `Neon2`[finite] into the (finite-sum version of) `CDHS` method [8], we have[3]

**Theorem 6** (informal). `Neon2`[finite] *turns* `CDHS` *into a first-order method without hurting its performance: it finds an* $(\varepsilon, \delta)$-*approximate local minimum in* $T = \widetilde{O}\big(\frac{n}{\varepsilon^{1.5}} + \frac{n}{\delta^3} + \frac{n^{3/4}}{\varepsilon^{1.75}} + \frac{n^{3/4}}{\delta^{3.5}}\big)$ *stochastic gradient computations.*

*Remark* 1.1. All the cited works in Section 1.1 requires the objective to have (1) Lipschitz-continuous Hessian (a.k.a. second-order smoothness) and (2) Lipschitz-continuous gradient (a.k.a. Lipschitz smoothness). One can argue that (1) and (2) are both necessary for finding approximate local minima, but if only finding approximate stationary points, then only (2) is necessary. We shall formally discuss our assumptions in Section 2.

### 1.2 From Stationary Points to Local Minima

Given any first-order method that finds stationary points (such as `GD`, `SGD`, `SVRG` or `SCSG`), we can hope for using the NC-search routine to identify whether or not its output $x$ satisfies $\nabla^2 f(x) \succeq -\delta \mathbf{I}$. If so, then automatically $x$ becomes an $(\varepsilon, \delta)$-approximate local minima so we can terminate. If not, we can go into its negative curvature direction to further decrease the objective.

In the independent work of Xu and Yang [31], they applied their `Neon` method for NC-search, and thus turned `SGD` and `SCSG` into first-order methods finding approximate local minima. In this paper, we use `Neon2` instead. We show the following theorem:

---

[3]The original paper of `CDHS` only stated their algorithm in the deterministic setting, but is easily verifiable to work in the finite-sum setting, see discussions in [1].

| | algorithm | gradient complexity $T$ | Hessian-vector products | variance bound | Lip. smooth | 2$^{\text{nd}}$-order smooth |
|---|---|---|---|---|---|---|
| stationary | SGD (folklore) | $O\bigl(\frac{1}{\varepsilon^4}\bigr)$ | no | needed | needed | no |
| local minima | perturbed SGD [14] | $\widetilde{O}\bigl(\frac{\text{poly}(d)}{\varepsilon^4} + \frac{\text{poly}(d)}{\delta^{16}}\bigr)$ | no | needed | needed | needed |
| | Neon+SGD [31] | $\widetilde{O}\bigl(\frac{1}{\varepsilon^4} + \frac{1}{\delta^7}\bigr)$ | no | needed | needed | needed |
| | Neon2+SGD | $\widetilde{O}\bigl(\frac{1}{\varepsilon^4} + \frac{1}{\varepsilon^2\delta^3} + \frac{1}{\delta^5}\bigr)$ | no | needed | needed | needed |
| stationary | SCSG [19] | $O\bigl(\frac{1}{\varepsilon^{10/3}}\bigr)$ | no | needed | needed | no |
| local minima | Neon+SCSG [31] | $O\bigl(\frac{1}{\varepsilon^{10/3}} + \frac{1}{\varepsilon^2\delta^3} + \frac{1}{\delta^6}\bigr)$ | no | needed | needed | needed |
| | Neon2+SCSG | $O\bigl(\frac{1}{\varepsilon^{10/3}} + \frac{1}{\varepsilon^2\delta^3} + \frac{1}{\delta^5}\bigr)$ | no | needed | needed | needed |
| local minima | Natasha2 [2] | $\widetilde{O}\bigl(\frac{1}{\varepsilon^{3.25}} + \frac{1}{\varepsilon^3\delta} + \frac{1}{\delta^5}\bigr)$ | needed | needed | needed | needed |
| | Neon+Natasha2 [31] | $\widetilde{O}\bigl(\frac{1}{\varepsilon^{3.25}} + \frac{1}{\varepsilon^3\delta} + \frac{1}{\delta^6}\bigr)$ | no | needed | needed | needed |
| | Neon2+Natasha2 | $\widetilde{O}\bigl(\frac{1}{\varepsilon^{3.25}} + \frac{1}{\varepsilon^3\delta} + \frac{1}{\delta^5}\bigr)$ | no | needed | needed | needed |
| ↑ online methods ↑ | | | ↓ offline methods ↓ | | | |
| stationary | GD (folklore) | $O\bigl(\frac{n}{\varepsilon^2}\bigr)$ | no | no | needed | no |
| local minima | perturbed GD [16] | $\widetilde{O}\bigl(\frac{n}{\varepsilon^2} + \frac{n}{\delta^4}\bigr)$ | no | no | needed | needed |
| | Neon2+GD | $\widetilde{O}\bigl(\frac{n}{\varepsilon^2} + \frac{n}{\delta^{3.5}}\bigr)$ | no | no | needed | needed |
| stationary | SVRG [4, 24] | $O\bigl(\frac{n^{2/3}}{\varepsilon^2} + n\bigr)$ | no | no | needed | no |
| local minima | Reddi et al. [25] | $\widetilde{O}\bigl(\frac{n^{2/3}}{\varepsilon^2} + \frac{n}{\delta^3} + \frac{n^{3/4}}{\delta^{3.5}}\bigr)$ | needed | no | needed | needed |
| | Neon2+SVRG | | no | no | needed | needed |
| stationary | "guilty" [9] | $\widetilde{O}\bigl(\frac{n}{\varepsilon^{1.75}}\bigr)$ | no | no | needed | needed |
| local minima | FastCubic [1] CDHS [8] | $\widetilde{O}\bigl(\frac{n}{\varepsilon^{1.5}} + \frac{n}{\delta^3} + \frac{n^{3/4}}{\varepsilon^{1.75}} + \frac{n^{3/4}}{\delta^{3.5}}\bigr)$ | needed | no | needed | needed |
| | Neon2+CDHS | $\widetilde{O}\bigl(\frac{n}{\varepsilon^{1.5}} + \frac{n}{\delta^3} + \frac{n^{3/4}}{\varepsilon^{1.75}} + \frac{n^{3/4}}{\delta^{3.5}}\bigr)$ | no | no | needed | needed |

Table 1: Complexity for finding $\|\nabla f(x)\| \leq \varepsilon$ and $\nabla^2 f(x) \succeq -\delta \mathbf{I}$. Following tradition, in these complexity bounds, we assume variance and smoothness parameters as constants, and only show the dependency on $n, d, \varepsilon$.

**Remark 1.** Variance bounds is needed for online methods (first half of the table).

**Remark 2.** Lipschitz smoothness is needed for finding even approximate stationary points.

**Remark 3.** Second-order Lipschitz smoothness is needed for finding approximate local minima.

**Theorem 7** (informal). *To find an $(\varepsilon, \delta)$-approximate local minima,*

(a) Neon2+SGD *needs $T = \widetilde{O}\bigl(\frac{1}{\varepsilon^4} + \frac{1}{\varepsilon^2\delta^3} + \frac{1}{\delta^5}\bigr)$ stochastic gradients;*

(b) Neon2+SCSG *needs $T = \widetilde{O}\bigl(\frac{1}{\varepsilon^{10/3}} + \frac{1}{\varepsilon^2\delta^3} + \frac{1}{\delta^5}\bigr)$ stochastic gradients; and*

(c) Neon2+GD *needs $T = \widetilde{O}\bigl(\frac{n}{\varepsilon^2} + \frac{n}{\delta^{3.5}}\bigr)$ (so $\widetilde{O}\bigl(\frac{1}{\varepsilon^2} + \frac{1}{\delta^{3.5}}\bigr)$ full gradients).*

(d) Neon2+SVRG *needs $T = \widetilde{O}\bigl(\frac{n^{2/3}}{\varepsilon^2} + \frac{n}{\delta^3} + \frac{n^{3/4}}{\delta^{3.5}}\bigr)$ stochastic gradients.*

We make several comments as follows.

(a) One should compare Neon2+SGD to Ge et al. [14], where the authors showed SGD plus per-

turbation needs $T = \widetilde{O}(\mathsf{poly}(d)/\varepsilon^4)$ to find $(\varepsilon, \varepsilon^{1/4})$-approximate local minima. Their result is perhaps the first time that a theoretical guarantee for finding approximate local minima is given using first-order oracles.

To some extent, Theorem 7a is superior because we have (1) removed the $\mathsf{poly}(d)$ factor,[4] (2) achieved $T = \widetilde{O}(1/\varepsilon^4)$ as long as $\delta \geq \varepsilon^{2/3}$, and (3) a simpler analysis.

We also remark that, if using Neon instead of Neon2, one achieves a slightly worse complexity $T = \widetilde{O}\big(\frac{1}{\varepsilon^4} + \frac{1}{\delta^7}\big)$, see Figure 1(a) for a comparison.[5]

(b) Neon2+SCSG turns SCSG into a local-minimum finding algorithm. Again, if using Neon instead of Neon2, one gets a slightly worse complexity $T = \widetilde{O}\big(\frac{1}{\varepsilon^{10/3}} + \frac{1}{\varepsilon^2 \delta^3} + \frac{1}{\delta^6}\big)$, see Figure 1(b).

(c) One should compare Neon2+GD to Jin et al. [16], where the authors showed GD plus perturbation needs $\widetilde{O}(1/\varepsilon^2)$ full gradients to find $(\varepsilon, \varepsilon^{1/2})$-approximate local minima. This is perhaps the first time that one converts a stationary-point finding method (namely GD) into a local minimum-finding one, without hurting its performance.

To some extent, Theorem 7c is better because we use $\widetilde{O}(1/\varepsilon^2)$ full gradients as long as $\delta \geq \varepsilon^{4/7}$.

(d) Our result for Neon2+SVRG matches the recent work of Reddi et al. [25], but we do not need Hessian-vector product computations.

**Limitation.** There is some limitation to Neon2 (or Neon) to turn an algorithm finding (approximate) stationary points to that finding (approximate) local minima. Namely, given any algorithm $\mathcal{A}$, if the gradient complexity for $\mathcal{A}$ to find an $\varepsilon$-approximate stationary point is $T$, then after this conversion, it finds $(\varepsilon, \delta)$-approximate local minima in a gradient complexity that is at least $T$. This is because the new algorithm, after combining Neon2 and $\mathcal{A}$, tries to alternatively find stationary points (using $\mathcal{A}$) and escape from saddle points (using Neon2). Therefore, it must pay at least complexity $T$.

In contrast, methods such as Natasha2 *swing by saddle points* instead of go to saddle points and escape. This has enabled it to achieve a smaller complexity $T = O(\varepsilon^{-3.25})$ for $\delta \geq \varepsilon^{1/4}$.

### 1.3 Roadmap

We introduce notions and formalize the problem in Section 2. We introduce Neon2 in the online, deterministic, and SVRG settings respectively in Section 3, Section 4 and Section 5. We apply Neon2 to SGD, GD, Natasha2, CDHS, SCSG and SVRG in Section 6. Most of the proofs are in the appendix.

## 2 Preliminaries

Throughout this paper, we denote by $\|\cdot\|$ the Euclidean norm. We use $i \in_R [n]$ to denote that $i$ is generated from $[n] = \{1, 2, \ldots, n\}$ uniformly at random. We denote by $\mathbb{I}[event]$ the indicator function of probabilistic events.

We denote by $\|\mathbf{A}\|_2$ the spectral norm of matrix $\mathbf{A}$. For symmetric matrices $\mathbf{A}$ and $\mathbf{B}$, we write $\mathbf{A} \succeq \mathbf{B}$ to indicate that $\mathbf{A} - \mathbf{B}$ is positive semidefinite (PSD). Therefore, $\mathbf{A} \succeq -\sigma \mathbf{I}$ if and only if all eigenvalues of $\mathbf{A}$ are no less than $-\sigma$. We denote by $\lambda_{\min}(\mathbf{A})$ and $\lambda_{\max}(\mathbf{A})$ the minimum and maximum eigenvalue of a symmetric matrix $\mathbf{A}$.

---

[4] We are aware that the original authors of [14] have a different proof to remove its $\mathsf{poly}(d)$ factor, but have not found it online at this moment.

[5] Their complexity is improvable to $\widetilde{O}\big(\frac{1}{\varepsilon^4} + \frac{1}{\delta^6}\big)$ with a slight change of the algorithm, but not beyond.

Recall some definitions on smoothness (for other equivalent definitions, see textbook [20])

**Definition 2.1.** *For a function $f\colon \mathbb{R}^d \to \mathbb{R}$,*
- *$f$ is $L$-Lipschitz smooth (or $L$-smooth for short) if*
$$\forall x, y \in \mathbb{R}^d, \ \|\nabla f(x) - \nabla f(y)\| \leq L\|x - y\|.$$
- *$f$ is second-order $L_2$-Lipschitz smooth (or $L_2$-second-order smooth for short) if*
$$\forall x, y \in \mathbb{R}^d, \ \|\nabla^2 f(x) - \nabla^2 f(y)\|_2 \leq L_2\|x - y\|.$$

The following fact says the variance of a random variable decreases by factor $m$ if we average $m$ independent copies.

**Fact 2.2.** *If $v_1, \ldots, v_n \in \mathbb{R}^d$ satisfy $\sum_{i=1}^n v_i = \vec{0}$, and $S$ is a non-empty, uniform random subset of $[n]$. Then*
$$\mathbb{E}\left[\left\|\tfrac{1}{|S|}\sum_{i\in S} v_i\right\|^2\right] = \tfrac{n-|S|}{(n-1)|S|} \cdot \tfrac{1}{n}\sum_{i\in[n]} \|v_i\|^2 \leq \tfrac{\mathbb{I}[|S|<n]}{|S|} \cdot \tfrac{1}{n}\sum_{i\in[n]} \|v_i\|^2 \ .$$

## 2.1 Problem and Assumptions

Throughout the paper we study the following minimization problem
$$\min_{x\in\mathbb{R}^d} \left\{ f(x) \stackrel{\text{def}}{=} \tfrac{1}{n}\sum_{i=1}^n f_i(x) \right\} \tag{2.1}$$

where both $f(\cdot)$ and each $f_i(\cdot)$ can be nonconvex. We wish to find $(\varepsilon, \delta)$-approximate local minima which are points $x$ satisfying
$$\|\nabla f(x)\| \leq \varepsilon \quad \text{and} \quad \nabla^2 f(x) \succeq -\delta \mathbf{I} \ .$$

We need the following three assumptions

- Each $f_i(x)$ is $L$-Lipschitz smooth.
- Each $f_i(x)$ is second-order $L_2$-Lipschitz smooth.

  (In fact, the gradient complexity of `Neon2` in this paper only depends polynomially on the second-order smoothness of $f(x)$ (rather than $f_i(x)$), and the time complexity depends logarithmically on the second-order smoothness of $f_i(x)$. To make notations simple, we decide to simply assume each $f_i(x)$ is $L_2$-second-order smooth.)

- Stochastic gradients have bounded variance: $\forall x \in \mathbb{R}^d\colon \quad \mathbb{E}_{i\in_R[n]} \|\nabla f(x) - \nabla f_i(x)\|^2 \leq \mathcal{V} \ .$

  (This assumption is needed only for *online* algorithms.)

# 3 Neon2 in the Online Setting

We propose `Neon2`<sup>online</sup> formally in Algorithm 1.

It repeatedly invokes `Neon2`<sup>online</sup><sub>weak</sub> in Algorithm 2, whose goal is to solve the NC-search problem with confidence $2/3$ only; then `Neon2`<sup>online</sup> invokes `Neon2`<sup>online</sup><sub>weak</sub> repeatedly for $\log(1/p)$ times to boost the confidence to $1 - p$. We prove the following theorem:

**Algorithm 1** $\texttt{Neon2}^{\textsf{online}}(f, x_0, \delta, p)$ ⋄ *for boosting confidence of $\texttt{Neon2}^{\textsf{online}}_{\textsf{weak}}$*

**Input:** Function $f(x) = \frac{1}{n}\sum_{i=1}^n f_i(x)$, vector $x_0$, negative curvature $\delta > 0$, confidence $p \in (0,1]$.
1: **for** $j = 1, 2, \cdots \Theta(\log 1/p)$ **do** ⋄ *boost the confidence*
2:     $v_j \leftarrow \texttt{Neon2}^{\textsf{online}}_{\textsf{weak}}(f, x_0, \delta)$;
3:     **if** $v_j \neq \bot$ **then**
4:         $m \leftarrow \Theta(\frac{L^2 \log 1/p}{\delta^2})$, $v' \leftarrow \Theta(\frac{\delta}{L_2})v$.
5:         Draw $i_1, \ldots, i_m \in_R [n]$.
6:         $z_j = \frac{1}{m\|v'\|_2^2} \sum_{j=1}^m (v')^\top \left(\nabla f_{i_j}(x_0 + v') - \nabla f_{i_j}(x_0)\right)$
7:         **if** $z_j \leq -3\delta/4$ **return** $v = v_j$
8:     **end if**
9: **end for**
10: **return** $v = \bot$.

---

**Algorithm 2** $\texttt{Neon2}^{\textsf{online}}_{\textsf{weak}}(f, x_0, \delta)$

1: $\eta \leftarrow \frac{\delta}{C_0^2 L^2 \log(100d)}$, $T \leftarrow \frac{C_0^2 \log(100d)}{\eta \delta}$,    ⋄ *for sufficiently large constant $C_0$*
2: $\xi \leftarrow$ Gaussian random vector with norm $\sigma$.    ⋄ $\sigma \stackrel{\text{def}}{=} (100d)^{-3C_0} \frac{\eta^2 \delta^3}{L_2}$
3: $x_1 \leftarrow x_0 + \xi$.
4: **for** $t \leftarrow 1$ **to** $T$ **do**
5:     $x_{t+1} \leftarrow x_t - \eta \left(\nabla f_i(x_t) - \nabla f_i(x_0)\right)$ where $i \in_R [n]$.
6:     **if** $\|x_{t+1} - x_0\|_2 \geq r$ **then return** $v = \frac{x_{t+1} - x_0}{\|x_{t+1} - x_0\|_2}$    ⋄ $r \stackrel{\text{def}}{=} (100d)^{C_0} \sigma$
7: **end for**
8: **return** $v = \bot$;

---

> **Theorem 1** ($\texttt{Neon2}^{\textsf{online}}$). *Let $f(x) = \frac{1}{n}\sum_{i=1}^n f_i(x)$ where each $f_i$ is $L$-smooth and $L_2$-second-order smooth. For every point $x_0 \in \mathbb{R}^d$, every $\delta \in (0, L]$, every $p \in (0,1)$, the output*
> $$v = \texttt{Neon2}^{\textsf{online}}(f, x_0, \delta, p)$$
> *satisfies that, with probability at least $1 - p$:*
>
> 1. *If $v = \bot$, then $\nabla^2 f(x_0) \succeq -\delta \mathbf{I}$.*
> 2. *If $v \neq \bot$, then $\|v\|_2 = 1$ and $v^\top \nabla^2 f(x_0) v \leq -\frac{\delta}{2}$.*
>
> *Moreover, the total number of stochastic gradient evaluations $O\left(\frac{\log^2(d/p) L^2}{\delta^2}\right)$.*

The proof of Theorem 1 immediately follows from Lemma 3.1 and Lemma 3.2 below.

**Lemma 3.1** ($\texttt{Neon2}^{\textsf{online}}_{\textsf{weak}}$). *In the same setting as Theorem 1, the output $v = \texttt{Neon2}^{\textsf{online}}_{\textsf{weak}}(f, x_0, \delta)$ satisfies If $\lambda_{\min}(\nabla^2 f(x_0)) \leq -\delta$, then with probability at least $2/3$, $v \neq \bot$ and $v^\top \nabla^2 f(x_0) v \leq -\frac{3}{4}\delta$.*

*Proof sketch of Lemma 3.1.* We explain why $\texttt{Neon2}^{\textsf{online}}_{\textsf{weak}}$ works as follows. Starting from a randomly perturbed point $x_1 = x_0 + \xi$, it keeps updating $x_{t+1} \leftarrow x_t - \eta \left(\nabla f_i(x_t) - \nabla f_i(x_0)\right)$ for some random index $i \in [n]$, and stops either when $T$ iterations are reached, or when $\|x_{t+1} - x_0\|_2 > r$. Therefore, we have $\|x_t - x_0\|_2 \leq r$ throughout the iterations, and thus can approximate $\nabla^2 f_i(x_0)(x_t - x_0)$ using $\nabla f_i(x_t) - \nabla f_i(x_0)$, up to error $O(r^2)$. This is a small term based on our choice of $r$.

Ignoring the error term, our updates look like $x_{t+1} - x_0 = \left(\mathbf{I} - \eta \nabla^2 f_i(x_0)\right)(x_t - x_0)$. This is exactly the same as Oja's algorithm [22] which is known to approximately compute the minimum eigenvector of $\nabla^2 f(x_0) = \frac{1}{n}\sum_{i=1}^n f_i(x_0)$. Using the recent optimal convergence analysis of Oja's algorithm [7], one can conclude that after $T_1 = \Theta\big(\frac{\log\frac{r}{\sigma}}{\eta\lambda}\big)$ iterations, where $\lambda = \max\{0, -\lambda_{\min}(\nabla^2 f(x_0))\}$, then we not only have that $\|x_{t+1} - x_0\|_2$ is blown up, but also it aligns well with the minimum eigenvector of $\nabla^2 f(x_0)$. In other words, if $\lambda \geq \delta$, then the algorithm must stop before $T$.

Finally, one has to carefully argue that the error does not blow up in this iterative process. We defer the proof details to Appendix A.3. □

Our Lemma 3.2 below tells us we can verify if the output $v$ of $\mathtt{Neon2}^{\mathsf{online}}_{\mathsf{weak}}$ is indeed correct (up to additive $\frac{\delta}{4}$), so we can boost the success probability to $1 - p$.

**Lemma 3.2** (verification). *In the same setting as Theorem 1, let vectors $x, v \in \mathbb{R}^d$. If $i_1, \ldots, i_m \in_R [n]$ and define*
$$z = \tfrac{1}{m}\sum_{j=1}^m v^\top(\nabla f_{i_j}(x+v) - \nabla f_{i_j}(x))$$
*Then, if $\|v\| \leq \frac{\delta}{8L_2}$ and $m = \Theta(\frac{L^2 \log 1/p}{\delta^2})$, with probability at least $1 - p$,*
$$\left|\tfrac{z}{\|v\|_2^2} - \tfrac{v^\top \nabla^2 f(x) v}{\|v\|_2^2}\right| \leq \tfrac{\delta}{4} \ .$$

The simple proof of Lemma 3.2 can be found in Section A.4.

## 4 Neon2 in the Deterministic Setting

---
**Algorithm 3** $\mathtt{Neon2}^{\mathsf{det}}(f, x_0, \delta, p)$
---
**Input:** A function $f$, vector $x_0$, negative curvature target $\delta > 0$, failure probability $p \in (0,1]$.
1: $T \leftarrow \frac{C_1^2 \log(d/p)\sqrt{L}}{\sqrt{\delta}}$. ⋄ *for sufficiently large constant $C_1$.*
2: $\xi \leftarrow$ Gaussian random vector with norm $\sigma$; ⋄ $\sigma \stackrel{\text{def}}{=} (d/p)^{-2C_1} \frac{\delta}{T^4 L_2}$
3: $x_1 \leftarrow x_0 + \xi$. $y_1 \leftarrow \xi, y_0 \leftarrow 0$
4: **for** $t \leftarrow 1$ **to** $T$ **do**
5:   $y_{t+1} = 2\mathcal{M}(y_t) - y_{t-1}$; ⋄ $\mathcal{M}(y) \stackrel{\text{def}}{=} -\frac{1}{L}(\nabla f(x_0 + y) - \nabla f(x_0)) + \big(1 - \frac{3\delta}{4L}\big)y$
6:   $x_{t+1} = x_0 + y_{t+1} - \mathcal{M}(y_t)$.
7:   **if** $\|x_{t+1} - x_0\|_2 \geq r$ **then return** $\frac{x_{t+1}-x_0}{\|x_{t+1}-x_0\|_2}$. ⋄ $r \stackrel{\text{def}}{=} (d/p)^{C_1}\sigma$
8: **end for**
9: **return** ⊥.
---

We propose $\mathtt{Neon2}^{\mathsf{det}}$ formally in Algorithm 3 and prove the following theorem:

> **Theorem 3** ($\mathtt{Neon2}^{\mathsf{det}}$). *Let $f(x)$ be a function that is $L$-smooth and $L_2$-second-order smooth. For every point $x_0 \in \mathbb{R}^d$, every $\delta > 0$, every $p \in (0,1]$, the output $v = \mathtt{Neon2}^{\mathsf{det}}(f, x_0, \delta, p)$ satisfies that, with probability at least $1 - p$:*
> 1. *If $v = \bot$, then $\nabla^2 f(x_0) \succeq -\delta \mathbf{I}$.*
> 2. *If $v \neq \bot$, then $\|v\|_2 = 1$ and $v^\top \nabla^2 f(x_0) v \leq -\frac{\delta}{2}$.*
>
> *Moreover, the total number full gradient evaluations is $O\big(\frac{\log^2(d/p)\sqrt{L}}{\sqrt{\delta}}\big)$.*

*Proof sketch of Theorem 3.* We explain the high-level intuition of $\mathtt{Neon2}^{\mathsf{det}}$ and the proof of Theorem 3 as follows. Define $\mathbf{M} = -\frac{1}{L}\nabla^2 f(x_0) + \big(1 - \frac{3\delta}{4L}\big)\mathbf{I}$. We immediately notice that

- all eigenvalues of $\nabla^2 f(x_0)$ in $\left[\frac{-3\delta}{4}, L\right]$ are mapped to the eigenvalues of $\mathbf{M}$ in $[-1, 1]$, and
- any eigenvalue of $\nabla^2 f(x_0)$ smaller than $-\delta$ is mapped to eigenvalue of $\mathbf{M}$ greater than $1 + \frac{\delta}{4L}$.

Therefore, as long as $T \geq \widetilde{\Omega}\left(\frac{L}{\delta}\right)$, if we compute $x_{T+1} = x_0 + \mathbf{M}^T \xi$ for some random vector $\xi$, by the theory of power method, $x_{T+1} - x_0$ must be a negative-curvature direction of $\nabla^2 f(x_0)$ with value $\leq \frac{1}{2}\delta$. There are *two issues* with this approach.

The first issue is that, the degree $T$ of this matrix polynomial $\mathbf{M}^T$ can be reduced to $T = \widetilde{\Omega}\left(\frac{\sqrt{L}}{\sqrt{\delta}}\right)$ if the so-called Chebyshev polynomial is used.

**Claim 4.1.** *Let $\mathcal{T}_t(x)$ be the t-th Chebyshev polynomial of the first kind, defined as:*

$$\mathcal{T}_0(x) \stackrel{\mathrm{def}}{=} 1, \qquad \mathcal{T}_1(x) \stackrel{\mathrm{def}}{=} x, \qquad \mathcal{T}_{n+1}(x) \stackrel{\mathrm{def}}{=} 2x \cdot \mathcal{T}_n(x) - \mathcal{T}_{n-1}(x)$$

*then $\mathcal{T}_t(x)$ satisfies (see Trefethen [30]):*

$$\mathcal{T}_t(x) = \begin{cases} \cos(n \arccos(x)) \in [-1, 1] & \text{if } x \in [-1, 1]; \\ \frac{1}{2}\left[\left(x - \sqrt{x^2 - 1}\right)^n + \left(x + \sqrt{x^2 - 1}\right)^n\right] & \text{if } x > 1. \end{cases}$$

Since $\mathcal{T}_t(x)$ stays between $[-1, 1]$ when $x \in [-1, 1]$, and grows to $\approx (1 + \sqrt{x^2 - 1})^t$ for $x \geq 1$, we can use $\mathcal{T}_T(\mathbf{M})$ in replacement of $\mathbf{M}^T$. Then, any eigenvalue of $\mathbf{M}$ that is above $1 + \frac{\delta}{4L}$ shall grow in a speed like $(1 + \sqrt{\delta/L})^T$, so it suffices to choose $T \geq \widetilde{\Omega}\left(\frac{\sqrt{L}}{\sqrt{\sigma}}\right)$. This is quadratically faster than applying the power method, so in `Neon2det` we wish to compute $x_{t+1} \approx x_0 + \mathcal{T}_t(\mathbf{M})\xi$.

The second issue is that, since we cannot compute Hessian-vector products, we have to use the gradient difference to approximate it; that is, we can only use $\mathcal{M}(y)$ to approximate $\mathbf{M}y$ where

$$\mathcal{M}(y) \stackrel{\mathrm{def}}{=} -\frac{1}{L}\left(\nabla f(x_0 + y) - \nabla f(x_0)\right) + \left(1 - \frac{3\delta}{4L}\right)y .$$

How does error propagate if we compute $\mathcal{T}_t(\mathbf{M})\xi$ by replacing $\mathbf{M}$ with $\mathcal{M}$? Note that this is a very non-trivial question, because the coefficients of the polynomial $\mathcal{T}_t(x)$ is as large as $2^{O(t)}$.

It turns out, the way that error propagates depends on how the Chebyshev polynomial is calculated. If the so-called backward recurrence formula is used, namely,

$$y_0 = 0, \quad y_1 = \xi, \quad y_t = 2\mathcal{M}(y_{t-1}) - y_{t-2}$$

and setting $x_{T+1} = x_0 + y_{T+1} - \mathcal{M}(y_T)$, then this $x_{T+1}$ is sufficiently close to the exact value $x_0 + \mathcal{T}_t(\mathbf{M})\xi$. This is known as the stability theory of computing Chebyshev polynomials, and is proved in [6].

We defer all the proof details to Appendix B.2. □

## 5 Neon2 in the Finite-Sum Setting

Let us recall how the shift-and-invert (SI) approach [26] solves the minimum eigenvector problem. Given matrix $\mathbf{A} = \nabla^2 f(x_0) \in \mathbb{R}^{d \times d}$ and suppose its eigenvalues are $-L \leq \lambda_1 \leq \cdots \leq \lambda_d \leq L$. At a high level, the SI approach

- chooses $\lambda = \delta - \lambda_1$ ,[6]
- defines positive definite matrix $\mathbf{B} = (\lambda \mathbf{I} + \mathbf{A})^{-1}$, and

---
[6]The precise SI approach needs to binary search $\lambda$ because $\lambda_1$ is unknown.

- applies power method for a logarithmic number of rounds to $\mathbf{B}$ to find its approximate maximum eigenvector $v$.[7]

One can show that this unit vector $v$ satisfies $\lambda_1 \leq v^\top \mathbf{A} v \leq \lambda_1 + O(\delta)$ [13].

To apply power method to $\mathbf{B}$, one needs to compute matrix inversion $\mathbf{B}y = (\lambda \mathbf{I} + \mathbf{A})^{-1} y$ for arbitrary vectors $y \in \mathbb{R}^d$. The stability of SI ensures that it suffices to compute $\mathbf{B}y$ to some sufficiently high accuracy.[8]

One efficient way to compute $\mathbf{B}y$ to such high accuracy is by expressing $\mathbf{A}$ in a finite-sum form and then adopt convex optimization [13]. We call this approach *finite-sum SI*. Consider a *convex* quadratic function that is of a *finite sum of non-convex* functions:

$$g(z) \stackrel{\text{def}}{=} \frac{1}{2} z^\top (\lambda \mathbf{I} + \mathbf{A}) z + y^\top z = \frac{1}{n} \sum_{i=1}^n \left( \frac{1}{2} z^\top (\lambda \mathbf{I} + \nabla^2 f_i(x_0)) z + y^\top z \right) =: \frac{1}{n} \sum_{i=1}^n g_i(z) \ .$$

Now, computing $\mathbf{B}y$ is equivalent to minimizing $g(z)$, and one can use a stochastic first-order method to minimize it.

One such method is `KatyushaX`, which directly accelerates the so-called SVRG method using momentum, and finds $z$ using $\widetilde{O}(n + n^{3/4}\sqrt{L/\delta})$ computations of stochastic gradients.[9] Whenever a stochastic gradient $\nabla g_i(z) = (\lambda \mathbf{I} + \nabla^2 f_i(x_0)) z + y$ is needed at some point $z \in \mathbb{R}^d$ for some random $i \in [n]$, instead of evaluating it exactly (which require a Hessian-vector product), we use $\nabla f_i(x_0 + z) - \nabla f_i(x_0)$ to approximate $\nabla^2 f_i(x_0) \cdot z$. We call this method `Neon2`$^{\mathsf{finite}}$.

Of course, one needs to show that `KatyushaX` is stable to noise. Using similar techniques as the previous two sections, one can show that the error term is proportional to $O(\|z\|_2^2)$, and thus as long as we bound the norm of $z$ is bounded (just like we did in the previous two sections), this should not affect the performance of the algorithm. We decide to ignore the detailed theoretical proof of this result, because it will complicate this paper.

**Theorem 5** (`Neon2`$^{\mathsf{finite}}$)**.** *Let $f(x) = \frac{1}{n} \sum_{i=1}^n f_i(x)$ where each $f_i$ is $L$-smooth and $L_2$-second-order smooth. For every point $x_0 \in \mathbb{R}^d$, every $\delta > 0$, every $p \in (0, 1]$, the output $v = $ `Neon2`$^{\mathsf{finite}}(f, x_0, \delta, p)$ satisfies that, with probability at least $1 - p$:*

1. *If $v = \bot$, then $\nabla^2 f(x_0) \succeq -\delta \mathbf{I}$.*
2. *If $v \neq \bot$, then $\|v\|_2 = 1$ and $v^\top \nabla^2 f(x_0) v \leq -\frac{\delta}{2}$.*

*Moreover, the total number stochastic gradient evaluations is $\widetilde{O}\big(n + \frac{n^{3/4}\sqrt{L}}{\sqrt{\delta}}\big)$, where the $\widetilde{O}$ notion hides logarithmic factors in $d, 1/p$ and $L/\delta$.*

## 6 Applications of Neon2

We show how `Neon2` can be applied to existing algorithms such as `SGD`, `GD`, `SCSG`, `SVRG`, `Natasha2`, `CDHS`. Unfortunately, we are unaware of a generic statement for applying `Neon2` to *any* algorithm.

---

[7] More precisely, applying power method for $O(\log(d/p))$ rounds, one can find a unit vector $v$ such that $v^\top \mathbf{B} v \geq \frac{9}{10} \lambda_{\max}(\mathbf{B})$ with probability at least $1 - p$. One can also prove that this vector $v$ satisfies $\lambda_1 \leq v^\top \mathbf{A} v \leq \lambda_1 + O(\delta)$.

[8] More precisely, if suffices to compute $w \in \mathbb{R}^d$ so that $\|w - \mathbf{B}y\| \leq \varepsilon \|y\|$, in a time complexity that polynomially depends on $\log \frac{1}{\varepsilon}$ [5, 13].

[9] Shalev-Shwartz [29] first discovered that one can apply SVRG to minimize sum-of-nonconvex functions. It was also observed that applying APPA/Catalyst reductions to SVRG one can achieve accelerated convergence rates [13, 29], and this approach is commonly known as AccSVRG. However, AccSVRG requires some careful parameter tuning of its inner loops, and thus is a logarithmic-factor slower than `KatyushaX` and also less practical [3].

**Algorithm 4** $\texttt{Neon2+SGD}(f, x_0, p, \varepsilon, \delta)$

**Input:** function $f(\cdot)$, starting vector $x_0$, confidence $p \in (0, 1)$, $\varepsilon > 0$ and $\delta > 0$.
1: $K \leftarrow O\big(\frac{L_2^2 \Delta_f}{\delta^3} + \frac{L \Delta_f}{\varepsilon^2}\big)$; ⋄ $\Delta_f$ *is any upper bound on* $f(x_0) - \min_x\{f(x)\}$
2: **for** $t \leftarrow 0$ **to** $K - 1$ **do**
3:     $S \leftarrow$ a uniform random subset of $[n]$ with cardinality $|S| = B \stackrel{\text{def}}{=} \max\{\frac{8\mathcal{V}}{\varepsilon^2}, 1\}$;
4:     $x_{t+1/2} \leftarrow x_t - \frac{1}{L|S|} \sum_{i \in S} \nabla f_i(x_t)$;
5:     **if** $\|\nabla f(x_t)\| \geq \frac{\varepsilon}{2}$ **then** ⋄ *estimate* $\|\nabla f(x_t)\|$ *using* $O(\varepsilon^{-2} \mathcal{V} \log(K/p))$ *stochastic gradients*
6:         $x_{t+1} \leftarrow x_{t+1/2}$;
7:     **else** ⋄ *necessarily* $\|\nabla f(x_t)\| \leq \varepsilon$
8:         $v \leftarrow \texttt{Neon2}^{\text{online}}(x_t, \delta, \frac{p}{2K})$;
9:         **if** $v = \perp$ **then return** $x_t$; ⋄ *necessarily* $\nabla^2 f(x_t) \succeq -\delta \mathbf{I}$
10:        **else** $x_{t+1} \leftarrow x_t \pm \frac{\delta}{L_2} v$; ⋄ *necessarily* $v^\top \nabla^2 f(x_t) v \leq -\delta/2$
11:     **end if**
12: **end for**
13: will not reach this line (with probability $\geq 1 - p$).

---

**Algorithm 5** $\texttt{Neon2+GD}(f, x_0, p, \varepsilon, \delta)$

**Input:** function $f(\cdot)$, starting vector $x_0$, confidence $p \in (0, 1)$, $\varepsilon > 0$ and $\delta > 0$.
1: $K \leftarrow O\big(\frac{L_2^2 \Delta_f}{\delta^3} + \frac{L \Delta_f}{\varepsilon^2}\big)$; ⋄ $\Delta_f$ *is any upper bound on* $f(x_0) - \min_x\{f(x)\}$
2: **for** $t \leftarrow 0$ **to** $K - 1$ **do**
3:     $x_{t+1/2} \leftarrow x_t - \frac{1}{L} \nabla f(x_t)$;
4:     **if** $\|\nabla f(x_t)\| \geq \frac{\varepsilon}{2}$ **then**
5:         $x_{t+1} \leftarrow x_{t+1/2}$;
6:     **else**
7:         $v \leftarrow \texttt{Neon2}^{\text{det}}(x_t, \delta, \frac{p}{2K})$;
8:         **if** $v = \perp$ **then return** $x_t$; ⋄ *necessarily* $\nabla^2 f(x_t) \succeq -\delta \mathbf{I}$
9:         **else** $x_{t+1} \leftarrow x_t \pm \frac{\delta}{L_2} v$; ⋄ *necessarily* $v^\top \nabla^2 f(x_t) v \leq -\delta/2$
10:     **end if**
11: **end for**
12: will not reach this line (with probability $\geq 1 - p$).

---

Therefore, we have to prove them individually.[10]

Throughout this section, we assume that some starting vector $x_0 \in \mathbb{R}^d$ and upper bound $\Delta_f$ is given to the algorithm, and it satisfies $f(x_0) - \min_x\{f(x)\} \leq \Delta_f$. This is only for the purpose of proving theoretical bounds. In practice, because $\Delta_f$ only appears in specifying the number of iterations, can just run enough number of iterations and then halt the algorithm, without the necessity of knowing $\Delta_f$.

## 6.1 Applying Neon2 to SGD and GD

To apply $\texttt{Neon2}$ to turn $\texttt{SGD}$ into an algorithm finding approximate local minima, we propose the following process $\texttt{Neon2+SGD}$ (see Algorithm 4). In each iteration $t$, we first apply $\texttt{SGD}$ with mini-batch size $O(\frac{1}{\varepsilon^2})$ (see Line 4). Then, if $\texttt{SGD}$ finds a point with small gradient, we apply $\texttt{Neon2}^{\text{online}}$ to decide if it has a negative curvature, if so, then we move in the direction of the negative curvature (see Line 10). We have the following theorem:

> **Theorem 7a.** *With probability at least $1 - p$, Neon2+SGD outputs an $(\varepsilon, \delta)$-approximate local minimum in gradient complexity $T = \widetilde{O}\Big(\big(\frac{\mathcal{V}}{\varepsilon^2} + 1\big)\big(\frac{L_2^2 \Delta_f}{\delta^3} + \frac{L \Delta_f}{\varepsilon^2}\big) + \frac{L^2}{\delta^2} \frac{L_2^2 \Delta_f}{\delta^3}\Big)$.*

---

[10]This is because stationary-point finding algorithms have somewhat different guarantees. For instance, in mini-batch $\texttt{SGD}$ we have $f(x_t) - \mathbb{E}[f(x_{t+1})] \geq \Omega(\|\nabla f(x_t)\|^2)$ but in $\texttt{SCSG}$ we have $f(x_t) - \mathbb{E}[f(x_{t+1})] \geq \Omega(\mathbb{E}[\|\nabla f(x_{t+1})\|^2])$.

**Corollary 6.1.** *Treating $\Delta_f, \mathcal{V}, L, L_2$ as constants, we have $T = \widetilde{O}\big(\frac{1}{\varepsilon^4} + \frac{1}{\varepsilon^2 \delta^3} + \frac{1}{\delta^5}\big)$.*

One can similarly (and more easily) give an algorithm Neon2+GD, which is the same as Neon2+SGD except that the mini-batch SGD is replaced with a full gradient descent, and the use of Neon2$^\mathsf{online}$ is replaced with Neon2$^\mathsf{det}$. We have the following theorem:

> **Theorem 7c.** *With probability at least $1 - p$, Neon2+GD outputs an $(\varepsilon, \delta)$-approximate local minimum using $\widetilde{O}\Big(\frac{L\Delta_f}{\varepsilon^2} + \frac{L^{1/2}}{\delta^{1/2}} \frac{L_2^2 \Delta_f}{\delta^3}\Big)$ full gradient computations.*

We only prove Theorem 7a in Appendix C and the proof of Theorem 7c is only simpler.

## 6.2 Applying Neon2 to Natasha2 and CDHS

The recent results Carmon et al. [8] (that we refer to CDHS) and Natasha2 [2] are both Hessian-free methods, where the only Hessian-vector computations come from the NC-search process. Therefore, by replacing their NC-search with Neon2, we can turn them into pure first-order methods without Hessian-vector products.

We state the following two theorems where the proofs are exactly the same as the papers [8] and [2]. We directly state them by assuming $\Delta_f, \mathcal{V}, L, L_2$ are constants, to simplify our notions.

**Theorem 2.** *One can replace Oja's algorithm with Neon2$^\mathsf{online}$ in Natasha2 without hurting its performance, turning it into a first-order stochastic method.*

*Treating $\Delta_f, \mathcal{V}, L, L_2$ as constants, Natasha2 finds an $(\varepsilon, \delta)$-approximate local minimum in $T = \widetilde{O}\big(\frac{1}{\varepsilon^{3.25}} + \frac{1}{\varepsilon^3 \delta} + \frac{1}{\delta^5}\big)$ stochastic gradient computations.*

**Theorem 4.** *One can replace Lanczos method with Neon2$^\mathsf{det}$ or Neon2$^\mathsf{finite}$ in CDHS without hurting it performance, turning it into a first-order method.*

*Treating $\Delta_f, L, L_2$ as constants, CDHS finds an $(\varepsilon, \delta)$-approximate local minimum in either $\widetilde{O}\big(\frac{1}{\varepsilon^{1.75}} + \frac{1}{\delta^{3.5}}\big)$ full gradient computations (if Neon2$^\mathsf{det}$ is used) or in $T = \widetilde{O}\big(\frac{n}{\varepsilon^{1.5}} + \frac{n}{\delta^3} + \frac{n^{3/4}}{\varepsilon^{1.75}} + \frac{n^{3/4}}{\delta^{3.5}}\big)$ stochastic gradient computations (if Neon2$^\mathsf{finite}$ is used).*

## 6.3 Applying Neon2 to SCSG and SVRG

**Background.** We first recall the main idea of the SVRG method for non-convex optimization [4, 24]. It is an *offline* method but is what SCSG is built on. SVRG divides iterations into epochs, each of length $n$. It maintains a snapshot point $\widetilde{x}$ for each epoch, and computes the full gradient $\nabla f(\widetilde{x})$ only for snapshots. Then, in each iteration $t$ at point $x_t$, SVRG defines gradient estimator $\widetilde{\nabla} f(x_t) \stackrel{\text{def}}{=} \nabla f_i(x_t) - \nabla f_i(\widetilde{x}) + \nabla f(\widetilde{x})$ which satisfies $\mathbb{E}_i[\widetilde{\nabla} f(x_t)] = \nabla f(x_t)$, and performs update $x_{t+1} \leftarrow x_t - \alpha \widetilde{\nabla} f(x_t)$ for learning rate $\alpha$.

The SCSG method of Lei et al. [19] proposed a simple fix to turn SVRG into an online method. They changed the epoch length of SVRG from $n$ to $B \approx 1/\varepsilon^2$, and then replaced the computation of $\nabla f(\widetilde{x})$ with $\frac{1}{|S|} \sum_{i \in S} \nabla f_i(\widetilde{x})$ where $S$ is a random subset of $[n]$ with cardinality $|S| = B$. To

**Algorithm 6** `Neon2+SCSG`$(f, x_0, \varepsilon, \delta)$

**Input:** function $f(\cdot)$, starting vector $x_0$, $\varepsilon > 0$ and $\delta > 0$.
1: $B \leftarrow \max\{1, \frac{48\mathcal{V}}{\varepsilon^2}\}$; $b \leftarrow \max\left\{1, \Theta(\frac{(\varepsilon^2+\mathcal{V})\varepsilon^4 L_2^6}{\delta^9 L^3})\right\}$;
2: **if** $b > B$ **then return** `Neon2+SGD`$(f, x_0, 2/3, \varepsilon, \delta)$;    ⋄ *for cleaner analysis purpose, see Remark 6.3*
3: $K \leftarrow \Theta\big(\frac{Lb^{1/3}\Delta_f}{\varepsilon^{4/3}\mathcal{V}^{1/3}}\big)$;    ⋄ $\Delta_f$ *is any upper bound on* $f(x_0) - \min_x\{f(x)\}$
4: **for** $t \leftarrow 0$ **to** $K-1$ **do**
5:    $x_{t+1/2} \leftarrow$ apply SCSG on $x_t$ for one epoch of size $B = \max\{\Theta(\mathcal{V}/\varepsilon^2), 1\}$;
6:    **if** $\|\nabla f(x_{t+1/2})\| \geq \frac{\varepsilon}{2}$ **then**    ⋄ *estimate* $\|\nabla f(x_t)\|$ *using* $O(\varepsilon^{-2}\mathcal{V}\log K)$ *stochastic gradients*
7:       $x_{t+1} \leftarrow x_{t+1/2}$;
8:    **else**    ⋄ *necessarily* $\|\nabla f(x_{t+1/2})\| \leq \varepsilon$
9:       $v \leftarrow$ `Neon2`$^{\text{online}}(f, x_{t+1/2}, \delta, 1/20K)$;
10:      **if** $v = \perp$ **then return** $x_{t+1/2}$;    ⋄ *necessarily* $\nabla^2 f(x_{t+1/2}) \succeq -\delta\mathbf{I}$
11:      **else** $x_{t+1} \leftarrow x_{t+1/2} \pm \frac{\delta}{L_2}v$;    ⋄ *necessarily* $v^\top \nabla^2 f(x_{t+1/2})v \leq -\delta/2$
12:   **end if**
13: **end for**
14: will not reach this line (with probability $\geq 2/3$).

---

make this approach even more general, they also analyzed SCSG in the mini-batch setting, with mini-batch size $b \in \{1, 2, \ldots, B\}$.[11] Their Theorem 3.1 [19] says that,

**Lemma 6.2** ([19]). *There exist constant $C > 1$ such that, if we run SCSG for an epoch of size $B$ (so using $O(B)$ stochastic gradients)[12] with mini-batch $b \in \{1, 2, \ldots, B\}$ starting from a point $x_t$ and moving to $x_t^+$, then*

$$\mathbb{E}\big[\|\nabla f(x_t^+)\|^2\big] \leq C \cdot L(b/B)^{1/3}\big(f(x_t) - \mathbb{E}[f(x_t^+)]\big) + \frac{6\mathcal{V}}{B} \ .$$

**Our Approach.** In principle, one can apply the same idea of Neon2+SGD on SCSG to turn it into an algorithm finding approximate local minima. Unfortunately, this is not quite possible because the left hand side of Lemma 6.2 is on $\mathbb{E}\big[\|\nabla f(x_t^+)\|^2\big]$, as opposed to $\|\nabla f(x_t)\|^2$ in SGD (see (C.1)). This means, instead of testing whether $x_t$ is a good local minimum (as we did in Neon2+SGD), this time we need to test whether $x_t^+$ is a good local minimum. This creates some extra difficulty so we need a different proof.

*Remark* 6.3. As for the parameters of SCSG, we simply use $B = \max\{1, \frac{48\mathcal{V}}{\varepsilon^2}\}$. However, choosing mini-batch size $b = 1$ does not necessarily give the best complexity, so a tradeoff $b = \Theta(\frac{(\varepsilon^2+\mathcal{V})\varepsilon^4 L_2^6}{\delta^9 L^3})\}$ is needed. (A similar tradeoff was also discovered by the authors of Neon [31].) Note that this quantity $b$ may be larger than $B$, and if this happens, SCSG becomes essentially equivalent to one iteration of SGD with mini-batch size $b$. Instead of analyzing this boundary case $b > B$ separately, we decide to simply run Neon2+SGD whenever $b > B$ happens, to simplify our proof.

We show the following theorem (proved in Appendix C)

---

[11]That is, they reduced the epoch length to $\frac{B}{b}$, and replaced $\nabla f_i(x_t) - \nabla f_i(\widetilde{x})$ with $\frac{1}{|S'|}\sum_{i \in S'}\big(\nabla f_i(x_t) - \nabla f_i(\widetilde{x})\big)$ for some $S'$ that is a random subset of $[n]$ with cardinality $|S'| = b$.

[12]We remark that Lei et al. [19] only showed that an epoch runs in an *expectation* of $O(B)$ stochastic gradients. We assume it is exact here to simplify proofs. One can for instance stop SCSG after $O(B \log \frac{1}{p})$ stochastic gradient computations, and then Lemma 6.2 will succeed with probability $\geq 1 - p$.

**Theorem 7b.** *With probability at least* $2/3$, Neon2+SCSG *outputs an* $(\varepsilon, \delta)$-*approximate local minimum in gradient complexity* $T = \widetilde{O}\Big(\big(\frac{L\Delta_f}{\varepsilon^{4/3}\mathcal{V}^{1/3}} + \frac{L_2^2\Delta_f}{\delta^3}\big)\big(\frac{\mathcal{V}}{\varepsilon^2} + \frac{L^2}{\delta^2}\big) + \frac{L\Delta_f}{\varepsilon^2}\frac{L^2}{\delta^2}\Big).$

(To provide the simplest proof, we have shown Theorem 7b only with probability $2/3$. One can for instance boost the confidence to $1-p$ by running $\log \frac{1}{p}$ copies of Neon2+SCSG.)

**Corollary 6.4.** *Treating* $\Delta_f, \mathcal{V}, L, L_2$ *as constants, we have* $T = \widetilde{O}\big(\frac{1}{\varepsilon^{10/3}} + \frac{1}{\varepsilon^2\delta^3} + \frac{1}{\delta^5}\big).$

As for SVRG, it is an offline method and its one-epoch lemma looks like[13]
$$\mathbb{E}\big[\|\nabla f(x_t^+)\|^2\big] \leq C \cdot Ln^{1/3}\big(f(x_t) - \mathbb{E}[f(x_t^+)]\big) \ .$$

If one replaces the use of Lemma 6.2 with this new inequality, and replace the use of Neon2$^{\text{online}}$ with Neon2$^{\text{finite}}$, then we get the following theorem:

**Theorem 7d.** *With probability at least* $2/3$, Neon2+SVRG *outputs an* $(\varepsilon, \delta)$-*approximate local minimum in gradient complexity* $T = \widetilde{O}\Big(\frac{L\Delta_f n^{2/3}}{\varepsilon^2} + \frac{L_2^2\Delta_f}{\delta^3}\big(n + \frac{n^{3/4}\sqrt{L}}{\sqrt{\delta}}\big)\Big).$

For a clean presentation of this paper, we ignore the pseudocode and proof because they are only simpler than Neon2+SCSG.

## Acknowledgements

We would like to thank Tianbao Yang and Yi Xu for helpful feedbacks on this manuscript.

# APPENDIX

## A Proofs for Section 3: Neon2 in the Online Setting

### A.1 Auxiliary Lemmas

We use the following lemma to approximate hessian-vector products:

**Lemma A.1.** *If* $f(x)$ *is* $L_2$-*second-order smooth, then for every point* $x \in \mathbb{R}^d$ *and every vector* $v \in \mathbb{R}^d$, *we have:* $\|\nabla f(x+v) - \nabla f(x) - \nabla^2 f(x)v\|_2 \leq L_2\|v\|_2^2.$

*Proof of Lemma A.1.* We can write $\nabla f(x+v) - \nabla f(x) = \int_{t=0}^1 \nabla^2 f(x+tv)v dt$. Subtracting $\nabla^2 f(x)v$ we have:

$$\|\nabla f(x+v) - \nabla f(x) - \nabla^2 f(x)v\|_2 = \left\|\int_{t=0}^1 \big(\nabla^2 f(x+tv) - \nabla^2 f(x)\big)v dt\right\|_2$$
$$\leq \int_{t=0}^1 \|\nabla^2 f(x+tv) - \nabla^2 f(x)\|_2 \|v\|_2 dt \leq L_2\|v\|_2^2 \ . \qquad \square$$

We need the following auxiliary lemma on martingale concentration:

---
[13]There are at least three different variants of SVRG [4, 19, 24]. We have adopted the lemma of [19] for simplicity.

**Lemma A.2.** *Consider random events $\{\mathcal{F}_t\}_{t\geq 1}$ and random variables $x_1,\ldots,x_T \geq 0$ and $a_1,\ldots,a_T \in [-\rho, \rho]$ for $\rho \in [0, 1/2]$ where each $x_t$ and $a_t$ only depend on $\mathcal{F}_1,\ldots,\mathcal{F}_t$. Letting $x_0 = 0$ and suppose there exist constant $b \geq 0$ and $\mu > 0$ such that for every $t \geq 1$:*

$$x_t \leq x_{t-1}(1 - a_t) + b \quad \text{and} \quad \mathbb{E}[a_t \mid \mathcal{F}_1,\ldots,\mathcal{F}_{t-1}] \geq -\mu \ .$$

*Then, we have for every $p \in (0,1)$:* $\mathbf{Pr}\left[x_T \geq T \cdot b \cdot e^{\mu T + 2\rho\sqrt{T \log \frac{T}{p}}}\right] \leq p$ .

*Proof.* We have

$$\begin{aligned}
x_T &\leq (1 - a_T)x_{T-1} + b \leq (1 - a_T)\left((1 - a_{T-1})x_{T-2} + b\right) + b \\
&= (1 - a_T)(1 - a_{T-1})x_{T-2} + (1 - a_T)b + b \\
&\leq \cdots \leq \sum_{s=2}^{T} \prod_{t=s}^{T}(1 - a_s)b + b \ .
\end{aligned}$$

For every $s \in [T]$, we consider the random process $\{y_t^s\}_{t=s}^T$ define as

$$y_s^s = b \quad \text{and} \quad \text{for } t > s: \ y_t^s = (1 - a_t)y_{t-1}^s \ .$$

Therefore

$$\log y_t^s = \log(1 - a_t) + \log y_{t-1}^s$$

and we have $\log(1 - a_t) \in [-2\rho, \rho]$ and $\mathbb{E}[\log(1 - a_t) \mid \mathcal{F}_1, \cdots \mathcal{F}_{t-1}] \leq \mu$. Thus, applying Azuma-Hoeffding inequality on $\{\log y_t^s\}_{t=s}^T$, we have

$$\mathbf{Pr}\left[y_T^s \geq b \cdot e^{\mu T + 2\rho\sqrt{T \log \frac{T}{p}}}\right] \leq p/T \ .$$

Since $x_T \leq \sum_{s=1}^T y_T^s$, taking union bound over $s \in [T]$ we complete the proof. $\square$

We also need the following lemma to assist our analysis:

**Lemma A.3.** *Suppose $\mathbf{B}$ is a random symmetric matrix with $\|\mathbf{B}\|_2 \leq L$, such that $\mathbb{E}[\mathbf{B}] = -\mathbf{A}$ with $\lambda_{\max}(\mathbf{A}) = \lambda \geq 0$, then for every $\eta \in [0, 1/L]$, every unit vector $v$ with $v^\top \mathbf{A} v \geq 0$:*

$$v' = \frac{(\mathbf{I} - \eta\mathbf{B})v}{\|(\mathbf{I} - \eta\mathbf{B})v\|_2} \quad \text{satisfies} \quad \mathbb{E}_{\mathbf{B}}\left[(v')^\top \mathbf{A} v'\right] \geq v^\top \mathbf{A} v - O(\lambda \eta^2 L^2) \ .$$

*Proof of Lemma A.3.* We can directly calculate that

$$\begin{aligned}
\frac{v^\top(\mathbf{I} - \eta\mathbf{B})\mathbf{A}(\mathbf{I} - \eta\mathbf{B})v}{v^\top(\mathbf{I} - \eta\mathbf{B})(\mathbf{I} - \eta\mathbf{B})v} &= \frac{v^\top \mathbf{A} v - 2\eta v^\top \mathbf{B}\mathbf{A} v + \eta^2 v^\top \mathbf{B}\mathbf{A}\mathbf{B} v}{1 - 2\eta v^\top \mathbf{B} v + \eta^2 v^\top \mathbf{B}^2 v} \\
&\geq \frac{v^\top \mathbf{A} v - 2\eta v^\top \mathbf{B}\mathbf{A} v}{1 - 2\eta v^\top \mathbf{B} v + \eta^2 v^\top \mathbf{B}^2 v} \\
&\geq (v^\top \mathbf{A} v)(1 + 2\eta v^\top \mathbf{B} v) - 2\eta v^\top \mathbf{B}\mathbf{A} v - \lambda O(\eta^2 L^2)
\end{aligned}$$

Thus, since $\mathbb{E}[\mathbf{B}] = -\mathbf{A}$, we have:

$$\begin{aligned}
\mathbb{E}\left[\frac{v^\top(\mathbf{I} - \eta\mathbf{B})\mathbf{A}(\mathbf{I} - \eta\mathbf{B})v}{v^\top(\mathbf{I} - \eta\mathbf{B})(\mathbf{I} - \eta\mathbf{B})v}\right] &\geq v^\top \mathbf{A} v - 2\eta(v^\top \mathbf{A} v)^2 + 2\eta v^\top \mathbf{A}^2 v - \lambda O(\eta^2 L^2) \\
&\geq v^\top \mathbf{A} v - \lambda O(\eta^2 L^2) \ .
\end{aligned}$$

Above, we have used the Lieb-Thirring inequality:

$$(v^\top \mathbf{A} v)^2 = \mathbf{Tr}(\mathbf{A} v v^\top \mathbf{A} v v^\top) \leq \mathbf{Tr}(\mathbf{A}^2 v v^\top v v^\top) = v^\top \mathbf{A}^2 v \ . \quad \square$$

## A.2 Revisiting Oja's Method

We rewrite [7, Theorem 4] as Theorem A.4 below. The credit for discovering this theorem belongs to the prior work [7], but we apply some simple mathematical manipulations to make it useful for our paper. (We have not tried to improve the constants in the statement.)

**Theorem A.4** (Theorem 4 of [7] restated)**.** *There exists an absolute constant $C > 0$ such that the following holds: Suppose $\mathbf{B}_1, \mathbf{B}_2, \ldots, \mathbf{B}_t$ are i.i.d. random matrices with $\|\mathbf{B}_i\| \leq L$ and $\mathbb{E}[\mathbf{B}_i] = -\mathbf{A}$. Suppose also $\lambda_{\max}(\mathbf{A}) = \lambda \geq 0$. Let*

$$\forall i = 0, 1, \ldots, t: \quad v_{i+1} \stackrel{\text{def}}{=} (\mathbf{I} - \eta\mathbf{B}_i) \cdots (\mathbf{I} - \eta\mathbf{B}_1)\xi$$

*where $\xi$ is a random Gaussian vector with norm $\sigma$, and $\eta \in \left(0, \sqrt{\frac{1}{1350000 t L^2}}\right]$ is the learning rate. Then, with probability at least $99/100$:*

1. *Norm growth: $\|v_{t+1}\|_2 \geq \frac{1}{C}\left(e^{(\eta\lambda - 32\eta^2 L^2)t}\sigma/\sqrt{d}\right)$.*

2. *Negative curvature: $\frac{-v_{t+1}^\top \mathbf{A} v_{t+1}}{\|v_{t+1}\|_2^2} \leq -(1 - C\eta L)\lambda + C\left(\frac{\log d}{\eta t} + \sqrt{\frac{L^2}{t}} + \lambda\eta^2 L^2 t\right)$.*

Before giving the proof, we note a minor difference between Theorem A.4 and [7, Theorem 4]. In their statement, each matrix $\mathbf{B}_t$ is assumed (without loss of generality) to have all eigenvalues between 0 and 1. Therefore, to apply [7, Theorem 4] properly, we need a simple change of variable. This is for analysis purpose only, without the necessity of changing the algorithm.

Indeed, define $\mathbf{B}'_t = -\frac{\mathbf{B}_t}{2L} + \frac{\mathbf{I}}{2}$ so that $0 \preceq \mathbf{B}'_t \preceq \mathbf{I}$. We also let $\mathbf{A}' \stackrel{\text{def}}{=} \mathbb{E}[\mathbf{B}'_i]$ so we have $\lambda_{\max}(\mathbf{A}') = \frac{1}{2} + \frac{\lambda}{2L}$. It is an exercise to verify that we can rewrite

$$v_{t+1} = (1 - \eta L)^t w_{t+1} \quad \text{where} \quad w_{t+1} \stackrel{\text{def}}{=} (\mathbf{I} + \eta'\mathbf{B}'_t) \cdots (\mathbf{I} + \eta'\mathbf{B}'_1)\xi$$

for $\eta' \stackrel{\text{def}}{=} \frac{2\eta L}{1 - \eta L}$ and $w_1 = \xi$. This means, the $w_1, \ldots, w_{t+1}$ sequence satisfies the assumptions of [7, Theorem 4] so we can apply that theorem to $w_1, \ldots, w_{t+1}$.

*Proof of Theorem A.4.*

1. Denoting by $\mathbf{C} \stackrel{\text{def}}{=} (\mathbf{I} + \eta'\mathbf{B}'_t) \cdots (\mathbf{I} + \eta'\mathbf{B}'_1)$, the proof of [7, Theorem 4] —in particular, their Equation (I.4)— implies that as long as $(\eta')^2 t \leq \frac{1}{15000}$, we have with probability at least $199/200$,

$$\|\mathbf{C}\|_F \geq \frac{2}{3}e^{t\left(\eta'\left(\frac{1}{2} + \frac{\lambda}{2L}\right) - \eta'^2\left(\frac{1}{2} + \frac{\lambda}{2L}\right)^2\right)} \enspace.$$

Since $\xi$ is a random gaussian vector with norm $\sigma$, we know that for $w_{t+1} = \mathbf{C}\xi$, we have, with probability at least $199/200$:

$$\|w_{t+1}\|_2 \geq \frac{1}{200}\frac{\sigma}{\sqrt{d}}\|\mathbf{C}\|_F$$

Putting them together, we have with probability at least $99/100$,

$$\|w_{t+1}\|_2 \geq \Omega\left(e^{t\left(\eta'\left(\frac{1}{2} + \frac{\lambda}{2L}\right) - \eta'^2\left(\frac{1}{2} + \frac{\lambda}{2L}\right)^2\right)}\sigma/\sqrt{d}\right)$$

$$= \Omega\left(e^{t\left(\frac{2\eta L}{1 - \eta L}\left(\frac{1}{2} + \frac{\lambda}{2L}\right) - \left(\frac{2\eta L}{1 - \eta L}\right)^2\left(\frac{1}{2} + \frac{\lambda}{2L}\right)^2\right)}\sigma/\sqrt{d}\right)$$

$$\geq \Omega\left(e^{\frac{\eta\lambda t}{1 - \eta L} + \frac{\eta L t}{1 - \eta L} - 32\eta^2 L^2 t}\sigma/\sqrt{d}\right)$$

By $(1-x) \geq e^{-x/(1-x)}$ for $x \in [0,1]$, letting $x = \eta L$ we conclude:
$$\|v_{t+1}\|_2 = (1-\eta L)^t \|w_{t+1}\|_2 \geq \Omega\left(e^{\frac{\eta \lambda t}{1-\eta L} - 32\eta^2 L^2 t} \sigma/\sqrt{d}\right) \geq \Omega\left(e^{(\eta\lambda - 32\eta^2 L^2)t} \sigma/\sqrt{d}\right)$$

2. Using [7, Theorem 4], we can get that with probability at least 1999/2000, letting $\lambda' = \frac{1}{2} + \frac{\lambda}{2L}$, then as long as $(\eta')^2 t \leq \frac{1}{150000\lambda'}$, we have
$$\frac{1}{t} \sum_{i \in [t]} \left(\frac{-w_i^\top \mathbf{B}_i' w_i}{\|w_i\|_2^2}\right) \leq -(1-2\eta')\lambda' + O\left(\frac{\log d}{\eta' t}\right)$$

By $\mathbf{B}_i' = -\frac{\mathbf{B}_i}{2L} + \frac{\mathbf{I}}{2}$ and $\lambda' = \frac{1}{2} + \frac{\lambda}{2L}$, we get:
$$\frac{1}{t} \sum_{i \in [t]} \left(\frac{w_i^\top \left(\frac{\mathbf{B}_i}{2L} - \frac{\mathbf{I}}{2}\right) w_i}{\|w_i\|_2^2}\right) \leq -(1-2\eta')\left(\frac{1}{2} + \frac{\lambda}{2L}\right) + O\left(\frac{\log d}{\eta' t}\right)$$

Using $\eta' = \frac{2\eta L}{1-\eta L}$, by our choice of $\eta$, we can simplify above as:
$$\frac{1}{t} \sum_{i \in [t]} \left(\frac{w_i^\top \mathbf{B}_i w_i}{\|w_i\|_2^2}\right) \leq -(1-3\eta L)\lambda + O\left(\frac{\log d}{\eta t} + \eta L^2\right)$$

Since $v_i$ is a scaling of $w_i$, we also have:
$$\frac{1}{t} \sum_{i \in [t]} \left(\frac{v_i^\top \mathbf{B}_i v_i}{\|v_i\|_2^2}\right) \leq -(1-3\eta L)\lambda + O\left(\frac{\log d}{\eta t} + \eta L^2\right) \quad . \tag{A.1}$$

We next wish to replace $\mathbf{B}_i$ by $\mathbf{A}$ in the above inequality. For every $i \in [t]$, we have $v_i$ is independent of $\mathbf{B}_i$ and thus
$$\mathbb{E}\left[\frac{v_i^\top \mathbf{B}_i v_i}{\|v_i\|_2^2} \mid v_i\right] = -\frac{v_i^\top \mathbf{A} v_i}{\|v_i\|_2^2}$$

Let $\mathcal{F}_i = \{\xi, \mathbf{B}_1, \cdots, \mathbf{B}_i\}$ and $X_i = \sum_{j \leq i} \left(\frac{v_j^\top \mathbf{B}_j v_j}{\|v_j\|_2^2} - \frac{v_j^\top \mathbf{A} v_j}{\|v_j\|_2^2}\right)$, we know that $X_i$ is $\mathcal{F}_i$ measurable, $\mathbb{E}[X_i \mid \mathcal{F}_{i-1}] = X_{i-1}$, and $|X_i - X_{i-1}| \leq 2L$. By Azuma-Hoeffding inequality,
$$\forall i \in [t], \quad \forall \varepsilon > 0: \quad \mathbf{Pr}\left[|X_i| \geq \varepsilon\right] \leq 2e^{-\frac{\varepsilon^2}{8L^2(i+1)}} \quad .$$

Thus, with probability at least 1999/2000 it satisfies $X_t = O(\sqrt{t}L)$. Putting this back to (A.1), we have with probability at least 999/1000,
$$\frac{1}{t} \sum_{i=1}^t \left(\frac{-v_i^\top \mathbf{A} v_i}{\|v_i\|_2^2}\right) \leq -(1-3\eta L)\lambda + O\left(\frac{\log d}{\eta t} + \sqrt{\frac{L^2}{t}} + \eta L^2\right) \leq -(1-3\eta L)\lambda + O\left(\frac{\log d}{\eta t} + \sqrt{\frac{L^2}{t}}\right)$$

Thus, there exists $t_0 \in [t]$ such that $-\frac{v_{t_0}^\top \mathbf{A} v_{t_0}}{\|v_{t_0}\|_2^2} \leq -(1-3\eta L)\lambda + O\left(\frac{\log d}{\eta t} + \sqrt{\frac{L^2}{t}}\right)$. Conditioning on this $t_0$ —which happens with probability $\geq 999/1000$— we apply Lemma A.3 and it says
$$\mathbb{E}\left[\frac{-v_{t+1}^\top \mathbf{A} v_{t+1}}{\|v_{t+1}\|_2^2} \mid v_{t_0}\right] \leq \frac{-v_{t_0}^\top \mathbf{A} v_{t_0}}{\|v_{t_0}\|_2^2} + O(\lambda \eta^2 L^2 t) \leq -(1-3\eta)\lambda + O\left(\frac{\log d}{\eta t} + \sqrt{\frac{L^2}{t}}\right) + O(\lambda \eta^2 L^2 t)$$

Since $\frac{-v_{t+1}^\top \mathbf{A} v_{t+1}}{\|v_{t+1}\|_2^2} \geq -\lambda$, by Markov inequality we know that with probability at least 199/200,

$$\frac{-v_{t+1}^\top \mathbf{A} v_{t+1}}{\|v_{t+1}\|_2^2} \geq -(1 - 600\eta)\lambda + O\left(\frac{\log d}{\eta t} + \sqrt{\frac{L^2}{t}}\right) + O(\lambda \eta^2 L^2 t) \ . \qquad \square$$

## A.3 Proof of Lemma 3.1

We first recall the parameter choices from $\texttt{Neon2}_{\textsf{weak}}^{\textsf{online}}$:

$$\eta \leftarrow \frac{\delta}{C_0^2 L^2 \log(100d)} \ , \quad T = \frac{C_0^2 \log(100d)}{\eta \delta} \ , \quad \sigma \stackrel{\text{def}}{=} (100d)^{-3C_0} \frac{\eta^2 \delta^3}{L_2} \ , \text{and} \quad r \stackrel{\text{def}}{=} (100d)^{C_0} \sigma \quad \text{(A.2)}$$

where $C_0$ is sufficiently large constant. Also recall $\xi$ is a Gaussian random vector with norm $\sigma$,

$$x_1 = x_0 + \xi \quad \text{and} \quad x_{t+1} \leftarrow x_t - \eta \left(\nabla f_i(x_t) - \nabla f_i(x_0)\right) \quad \text{for } i \in_R [n].$$

If $\|x_{t+1} - x_0\|_2 \geq r$ at some $t$, $\texttt{Neon2}_{\textsf{weak}}^{\textsf{online}}$ outputs $v = \frac{x_{t+1} - x_0}{\|x_{t+1} - x_0\|_2}$; otherwise it outputs $v = \bot$.

*Proof of Lemma 3.1.* Let $i_t \in [n]$ be the random index $i$ chosen when computing $x_{t+1}$ from $x_t$ in Line 5 of $\texttt{Neon2}_{\textsf{weak}}^{\textsf{online}}$. By Lemma A.1, we know that for every $t \geq 1$,

$$\left\|\nabla f_{i_t}(x_t) - \nabla f_{i_t}(x_0) - \nabla^2 f_{i_t}(x_0)(x_t - x_0)\right\|_2 \leq L_2 \|x_t - x_0\|_2^2 \ .$$

Therefore, there exists error vector $\xi_t \in \mathbb{R}^d$ with $\|\xi_t\|_2 \leq L_2 \|x_t - x_0\|_2^2$ such that

$$(x_{t+1} - x_0) = (x_t - x_0) - \eta \nabla^2 f_{i_t}(x_0)(x_t - x_0) + \eta \xi_t \ .$$

For notational simplicity, let us denote by

$$z_t \stackrel{\text{def}}{=} x_t - x_0, \quad \mathbf{A}_t \stackrel{\text{def}}{=} \mathbf{B}_t + \mathbf{R}_t \quad \text{where} \quad \mathbf{B}_t \stackrel{\text{def}}{=} \nabla^2 f_{i_t}(x_0), \quad \mathbf{R}_t \stackrel{\text{def}}{=} -\frac{\xi_t z_t^\top}{\|z_t\|_2^2}$$

then it satisfies

$$z_{t+1} = z_t - \eta \mathbf{B}_t z_t + \eta \xi_t = (\mathbf{I} - \eta \mathbf{A}_t) z_t \ .$$

We have, as long as $\texttt{Neon2}_{\textsf{weak}}^{\textsf{online}}$ does not terminate, it satisfies

$$\|\mathbf{R}_t\|_2 \leq L_2 \|z_t\|_2 \leq L_2 \cdot r \ , \quad \|\mathbf{B}_t\|_2 \leq L \ , \text{and} \quad \|\mathbf{A}_t\|_2 \leq \|\mathbf{B}_t\|_2 + \|\mathbf{R}_t\|_2 \leq \|\mathbf{B}_t\|_2 + L_2 r \leq 2L \ .$$

Now, define $\Phi_t \stackrel{\text{def}}{=} z_{t+1} z_{t+1}^\top = (\mathbf{I} - \eta \mathbf{A}_t) \cdots (\mathbf{I} - \eta \mathbf{A}_1) \xi \xi^\top (\mathbf{I} - \eta \mathbf{A}_1) \cdots (\mathbf{I} - \eta \mathbf{A}_t)$ and

$$w_t \stackrel{\text{def}}{=} \frac{z_t}{\|z_t\|_2} = \frac{z_t}{(\mathbf{Tr}(\Phi_{t-1}))^{1/2}} \ .$$

Then, before $\texttt{Neon2}_{\textsf{weak}}^{\textsf{online}}$ terminates, we have:

$$\mathbf{Tr}(\Phi_t) = \mathbf{Tr}(\Phi_{t-1}) \left(1 - 2\eta w_t^\top \mathbf{A}_t w_t + \eta^2 w_t^\top \mathbf{A}_t^2 w_t\right)$$

$$\leq \mathbf{Tr}(\Phi_{t-1}) \left(1 - 2\eta w_t^\top \mathbf{A}_t w_t + 4\eta^2 L^2\right)$$

$$\leq \mathbf{Tr}(\Phi_{t-1}) \left(1 - 2\eta w_t^\top \mathbf{B}_t w_t + 2\eta \|\mathbf{R}_t\|_2 + 4\eta^2 L^2\right)$$

$$\stackrel{①}{\leq} \mathbf{Tr}(\Phi_{t-1}) \left(1 - 2\eta w_t^\top \mathbf{B}_t w_t + 8\eta^2 L^2\right) \ .$$

Above, ① is because our choice of parameters (A.2) satisfies $r \leq \eta \frac{L^2}{L_2}$. Therefore,

$$\log\left(\mathbf{Tr}(\Phi_t)\right) \leq \log\left(\mathbf{Tr}(\Phi_{t-1})\right) + \log\left(1 - 2\eta w_t^\top \mathbf{B}_t w_t + 8\eta^2 L^2\right) \ .$$

Let
$$\lambda = -\lambda_{\min}(\nabla^2 f(x_0)) = -\lambda_{\min}(\mathbb{E}_{\mathbf{B}_t}[\mathbf{B}_t]) \quad \text{and} \quad \mathbf{A} \stackrel{\text{def}}{=} \nabla^2 f(x_0) = \mathbb{E}_{\mathbf{B}_t}[\mathbf{B}_t] \ .$$

We know $w_t^\top \mathbf{B}_t w_t \in [-L, L]$ and $\mathbb{E}_{\mathbf{B}_t}\left[w_t^\top \mathbf{B}_t w_t \mid w_t\right] = w_t^\top \mathbf{A} w_t \geq -\lambda$ (recall the randomness of $\mathbf{B}_t$ is independent of $w_t$). Thus, by concavity of log, we also have

$$\mathbb{E}\left[\log\left(1 - 2\eta w_t^\top \mathbf{B}_t w_t + 8\eta^2 L^2\right) \mid w_t\right] \leq 2\eta\lambda + 8\eta^2 L^2$$

$$\log(1 - 2\eta w_t^\top \mathbf{B}_t w_t + 8\eta^2 L^2) \in [-2(2\eta L + 8\eta^2 L^2), 2\eta L + 8\eta^2 L^2] \in [-6\eta L, 3\eta L]$$

Denote by $p \stackrel{\text{def}}{=} 1/100$ for notation convenience. Applying Azuma-Hoeffding on $\log(\Phi_t)$ we have

$$\mathbf{Pr}\left[\log(\mathbf{Tr}[\Phi_t]) - \log(\mathbf{Tr}[\Phi_0]) \geq 2\eta\lambda t + 8\eta^2 L^2 t + 32\eta L \sqrt{t \log \frac{t}{p}}\right] \leq p/t \ .$$

In other words, with probability at least $1 - p$, $\text{Neon2}_{\text{weak}}^{\text{online}}$ will not terminate until $t \geq T_0$, where $T_0$ is given by:

$$T_0 = \frac{\log\left(\frac{r^2}{\sigma^2}\right)}{3} \min\left\{\frac{1}{2\eta\lambda}, \frac{1}{8\eta^2 L^2 \log \frac{1}{p} \log \frac{1}{\eta L}}\right\} \tag{A.3}$$

Indeed, the choice of $T_0$ in (A.3) implies that:

$$2\eta\lambda T_0 + 8\eta^2 L^2 T_0 + 32\eta L \sqrt{T_0 \log \frac{T_0}{p}} \leq \log\left(\frac{r^2}{\sigma^2}\right) \ .$$

Therefore, for $t \leq T_0$, we have

$$\mathbf{Pr}\left[\log(\mathbf{Tr}[\Phi_t]) - \log(\mathbf{Tr}[\Phi_0]) \geq \log \frac{r^2}{\sigma^2}\right] \leq p/T_0 \ .$$

By definition, $\mathbf{Tr}(\Phi_t) = \|x_{t+1} - x_0\|_2^2$ and $\mathbf{Tr}(\Phi_0) = \|\xi\|^2 = \sigma^2$. Thus, taking a union bound over $t \in [T_0]$, we know that with probability at least $1 - p$, for every $t \in [T_0]$, $\|x_{t+1} - x_0\|_2 < r$. Thus, the $\text{Neon2}_{\text{weak}}^{\text{online}}$ does not terminate before iteration $T_0$.

In the remainder of this proof, we want to show that when $\lambda \geq \delta$, our $\text{Neon2}_{\text{weak}}^{\text{online}}$ outputs a vector $v$ —with probability at least $2/3$— satisfying $v^\top \mathbf{A} v \leq -\frac{3}{4}\delta$.

We first note when $\lambda \geq \delta$, using our choice of $\eta$ in (A.2) and $T_0$ in (A.3), we have $T_0 \geq \frac{\log(r^2/\sigma^2)}{6\eta\lambda}$. Now, let "true" vector $v_{t+1} \stackrel{\text{def}}{=} (\mathbf{I} - \eta\mathbf{B}_t) \cdots (\mathbf{I} - \eta\mathbf{B}_1)\xi$ and we have

$$z_{t+1} - v_{t+1} = \prod_{s=1}^{t}(\mathbf{I} - \eta\mathbf{A}_s)\xi - \prod_{s=1}^{t}(\mathbf{I} - \eta\mathbf{B}_s)\xi = (\mathbf{I} - \eta\mathbf{B}_t)(z_t - v_t) - \eta\mathbf{R}_t z_t \ .$$

Thus, if we call $u_t \stackrel{\text{def}}{=} z_t - v_t$ with $u_1 = 0$, then, before $\text{Neon2}_{\text{weak}}^{\text{online}}$ stops, we have:

$$\|u_{t+1} - (\mathbf{I} - \eta\mathbf{B}_t)u_t\|_2 \leq \eta \|\mathbf{R}_t z_t\|_2 \leq \eta L_2 r^2$$

Using Young's inequality $\|a + b\|_2^2 \leq (1 + \beta)\|a\|_2^2 + \left(\frac{1}{\beta} + 1\right)\|b\|_2^2$ for every $\beta > 0$, we have:

$$\|u_{t+1}\|_2^2 \leq \left(1 + \eta^2 L^2\right) \|(\mathbf{I} - \eta\mathbf{B}_t)u_t\|_2^2 + 8\frac{L_2^2}{L^2} r^4 \leq \|u_t\|_2^2 \left(1 - 2\eta \frac{u_t \mathbf{B}_t u_t}{\|u_t\|_2^2} + 10\eta^2 L^2\right) + 8\left(\frac{L_2}{L}\right)^2 r^4 \ .$$

We can apply martingale concentration Lemma A.2 with $\rho = 2\eta L + 10\eta^2 L^2 \leq 2.5\eta L$ and

$\mu = 2\eta\lambda$. We conclude that

$$\mathbf{Pr}\left[\|u_t\|_2 \geq 16\frac{L_2}{L}r^2 t e^{\eta\lambda t + 8\eta L\sqrt{t\log\frac{t}{p}}}\right] \leq p \ . \tag{A.4}$$

We next apply Theorem A.4. Note that our choices of parameters $\eta$ and $T$ in (A.2) satisfy the presumption of Theorem A.4, so we conclude that for each $t \in [T]$, with probability at least $99/100$:

1. Norm growth: $\|v_t\|_2 \geq \frac{1}{C}\left(e^{(\eta\lambda - 32\eta^2 L^2)t}\sigma/\sqrt{d}\right)$.

2. Negative curvature: $\frac{v_t^\top \nabla^2 f(x_0) v_t}{\|v_t\|_2^2} \leq -(1 - C\eta L)\lambda + C\left(\frac{\log d}{\eta t} + \sqrt{\frac{L}{t}} + \lambda\eta^2 L^2 t\right)$.

Let us choose a fixed $T_1$ as (recall $\lambda \geq \delta$, $p = 1/100$ and $C_0$ is sufficiently large):

$$T_1 \stackrel{\text{def}}{=} \frac{\log\frac{2C\sqrt{d}r}{\sigma}}{\eta\lambda - 32\eta^2 L^2} = \frac{C_0(\log d/p) + \log(2C\sqrt{d})}{\eta\lambda - 32\eta^2 L^2} \leq \frac{2C_0 \cdot \log d/p}{\eta\lambda - 32\eta^2 L^2} \leq \frac{4C_0 \cdot \log d/p}{\eta\lambda} \leq \frac{4C_0 \cdot \log d/p}{\eta\delta} < T \ . \tag{A.5}$$

At $t = T_1$, by the "norm growth" property, we know that w.p. $\geq 99/100$,

$$\|v_{T_1}\|_2 \geq 2r = \frac{1}{C}\left(e^{(\eta\lambda - 32\eta^2 L^2)T_1}\sigma/\sqrt{d}\right) \ .$$

Combining this with (A.4), we have w.p. $\geq 98/100$,

$$\frac{\|u_{T_1}\|_2}{\|v_{T_1}\|_2} = \frac{16L_2 r^2 T_1 e^{\eta\lambda T_1 + 8\eta L\sqrt{T_1 \log\frac{T_1}{p}}}}{L \cdot 2r}$$

$$\leq 16C\left(\frac{\sqrt{d}L_2 r^2 T_1 e^{\eta\lambda T_1 + 8\eta L\sqrt{T_1 \log\frac{T_1}{p}}}}{L e^{(\eta\lambda - 32\eta^2 L^2)T_1}\sigma}\right)$$

$$\leq 16C\left(\frac{\sqrt{d}L_2 r^2 T_1}{L\sigma}e^{8\eta L\sqrt{T_1 \log\frac{T_1}{p}} + 32\eta^2 L^2 T_1}\right)$$

$$\stackrel{①}{\leq} 16C\left(\frac{\sqrt{d}L_2 r^2 T_1}{L\sigma}e^{16\sqrt{\log\frac{T_1}{p}}}\right) \stackrel{②}{\leq} 16C\left(\frac{\sqrt{d}L_2 r^2 T_1^2}{L\sigma p}\right) \stackrel{③}{\leq} \frac{\delta}{100L} \leq \frac{1}{100} \ . \tag{A.6}$$

Above, inequality ① uses our choice of parameters in (A.2) which gives $\eta^2 L^2 T_1 \leq \eta^2 L^2 T = 1$; inequality ② is because for sufficiently large $x$ we have $e^{16\sqrt{\log(x)}} < x$; and inequality ③ is due to

$$\frac{r^2}{\sigma} = (d/p)^{2C_0}\sigma = (d/p)^{-C_0}\frac{\eta^2\delta^3}{L_2}$$

and thus, for sufficiently large $C_0$ and $p = 1/100$, we know that: $(d/p)^{-C_0} \leq \frac{p}{d} \times \frac{1}{16C_0^3 \log^2\frac{d}{p}}$. This implies (here the last inequality uses (A.5))

$$\frac{r^2}{\sigma} \leq \frac{\eta^2\delta^3 p}{C_0 dL_2 \left(4C_0 \log(d/p)\right)^2} \leq \frac{\delta p}{C_0\sqrt{d}T_1^2 L_2} \ . \tag{A.7}$$

Thus, for sufficiently large $C_0 \geq 1600C$ we prove ③.

Putting together (A.6) and $\|v_{T_1}\|_2 \geq 2r$, we conclude with probability at least $97/100$, $\|z_{T_1}\|_2 = \|v_{T_1} + u_{T_1}\| \geq r$. This means $\texttt{Neon2}_{\text{weak}}^{\text{online}}$ must terminate within $T_1 \leq T$ iterations.

Moreover, recall w.p. $\geq 99/100$, $\texttt{Neon2}_{\text{weak}}^{\text{online}}$ will not terminate before iteration $T_0 \geq \frac{\log(r^2/\sigma^2)}{6\eta\lambda}$. Thus, w.p. $\geq 96/100$, $\texttt{Neon2}_{\text{weak}}^{\text{online}}$ terminates somewhere at $t \in [T_0, T_1]$ of termination.

Using the "negative curvature" property, we have w.p. at least $\geq 99/100$,

$$\frac{v_t^\top \mathbf{A} v_t}{\|v_t\|_2^2} \leq -(1 - C\eta L)\lambda + C\left(\frac{\log d}{\eta T_0} + \sqrt{\frac{L^2}{T_0}} + \lambda \eta^2 L^2 T_1\right) \ . \tag{A.8}$$

Since $T_0 \geq \frac{2\log(r/\sigma)}{6\eta\lambda} \geq \frac{C_0 \log(d/p)}{3\eta\lambda} \geq \frac{C_0 \log d}{3\eta\lambda}$, thus,

$$\frac{\log d}{\eta T_0} \leq \frac{3\lambda}{C_0} \tag{A.9}$$

Moreover, by our choice of $\eta$ we have $\eta L^2 \leq \frac{\delta}{C_0^2} \leq \frac{\lambda}{C_0^2}$ and thus

$$\frac{L^2}{T_0} \leq \frac{3\eta\lambda L^2}{C_0} \leq \frac{\lambda^2}{C_0} \quad \text{and} \quad \lambda\eta^2 L^2 T_1 \leq \lambda\eta^2 L^2 \frac{4C_0 \log(d/p)}{\eta\lambda} \leq \lambda \frac{4C_0 \log d}{C_0^2 \log d} \leq \frac{4\lambda}{C_0} \tag{A.10}$$

Putting (A.9) and (A.10) back to (A.8), we have for sufficiently large $C_0$, with probability at least $95/100$, $\texttt{Neon2}_{\textsf{weak}}^{\textsf{online}}$ terminates at $t \in [T_0, T_1]$ and

$$\frac{v_t^\top \mathbf{A} v_t}{\|v_t\|_2^2} \leq -\frac{15}{16}\lambda \leq -\frac{15}{16}\delta \ .$$

Finally, since we have $\|u_t + v_t\|_2 = \|z_t\|_2 \geq r$ (by the termination criterion), using (A.4) we have with probability at least $99/100$,

$$\frac{\|u_t\|_2}{\|u_t\|_2 + \|v_t\|_2} \leq \frac{\|u_t\|_2}{\|u_t + v_t\|_2} \leq \frac{16L_2 r^2 T_1 e^{\eta\lambda T_1 + 8\eta L\sqrt{T_1 \log \frac{T_1}{p}}}}{Lr} \leq \frac{\delta}{50L} \ .$$

Above, the last inequality is due to (A.6). This implies $\frac{\|u_t\|_2}{\|v_t\|_2} \leq \frac{\delta}{49L}$. In sum, we have with probability at least $94/100$:

$$\begin{aligned}
\frac{z_t^\top \mathbf{A} z_t}{\|z_t\|_2^2} &= \frac{\|v_t\|_2^2}{\|z_t\|_2^2} \cdot \frac{z_t^\top \mathbf{A} z_t}{\|v_t\|_2^2} \leq \frac{\|v_t\|_2^2}{\|z_t\|_2^2} \cdot \frac{v_t^\top \mathbf{A} v_t + 4L\|z_t - v_t\|_2 \|v_t\|_2}{\|v_t\|_2^2} \\
&\leq \frac{\|v_t\|_2^2}{\|z_t\|_2^2} \cdot \left(\frac{v_t^\top \mathbf{A} v_t}{\|v_t\|_2^2} + \frac{4L\|u_t\|_2}{\|v_t\|_2}\right) \leq \frac{\|v_t\|_2^2}{\|z_t\|_2^2} \cdot \left(-\frac{15}{16}\delta + \frac{4}{49}\delta\right) \leq -\frac{17\delta}{20} \frac{\|v_t\|_2^2}{\|z_t\|_2^2} \\
&\leq -\frac{17\delta}{20}\left(1 - \frac{\|u_t\|_2^2}{\|z_t\|_2^2}\right) \leq -\frac{17\delta}{20} \cdot \frac{49}{50} < -\frac{3}{4}\delta \ . \qquad \square
\end{aligned}$$

### A.4 Proof of Lemma 3.2

*Proof of Lemma 3.2.* By Lemma A.1, we know for every $i \in [n]$,

$$\left\|v^\top \left(\nabla f_i(x+v) - \nabla f_i(x) - \nabla^2 f_i(x)\right) v\right\|_2 \leq L_2 \|v\|_2^3 \ .$$

Letting $z_j = v^\top(\nabla f_{i_j}(x+v) - \nabla f_{i_j}(x))$, we know that $z_1, \cdots, z_m$ are i.i.d. random variables with $|z_j| \leq L\|v\|_2^2 + L_2\|v\|_2^3$. By Chernoff bound, we know that

$$\mathbf{Pr}\left[|z - \mathbb{E}[z]| \geq 2\left(L\|v\|_2^2 + L_2\|v\|_2^3\right) \sqrt{\frac{1}{m} \log \frac{1}{p}}\right] \leq p$$

Since we also have $\left|\mathbb{E}[z] - v^\top \nabla^2 f(x)v\right| \leq L_2\|v\|_2^3$ from Lemma A.1, we conclude that

$$\mathbf{Pr}\left[\left|\frac{z}{\|v\|_2^2} - \frac{v^\top \nabla^2 f(x)v}{\|v\|_2^2}\right| \leq 2\left(L + L_2\|v\|_2\right) \sqrt{\frac{1}{m}\log\frac{1}{p}} + L_2\|v\|_2\right] \geq 1 - p \ .$$

Plugging in our assumption on $\|v\|_2$ and our choice of $m$ finishes the proof. $\square$

# B Proofs for Section 4: Neon2 in the Deterministic Setting

## B.1 Stable Computation of Chebyshev Polynomials

We recall the following result from [6, Section 6.2] regarding one way to *stably* compute Chebyshev polynomials. Suppose we want to compute

$$\vec{s}_N \stackrel{\text{def}}{=} \sum_{k=0}^{N} \mathcal{T}_k(\mathbf{M})\vec{c}_k \in \mathbb{R}^d \quad \text{where } \mathbf{M} \in \mathbb{R}^{d \times d} \text{ is symmetric and each } \vec{c}_k \text{ is in } \mathbb{R}^d \ . \quad (B.1)$$

**Definition B.1** (inexact backward recurrence). *Let $\mathcal{M}$ be an approximate algorithm that satisfies $\|\mathcal{M}(u) - \mathbf{M}u\|_2 \leq \varepsilon \|u\|_2$ for every $u \in \mathbb{R}^d$. Then, define inexact backward recurrence to be*

$$\widehat{b}_{N+1} \stackrel{\text{def}}{=} 0, \quad \widehat{b}_N \stackrel{\text{def}}{=} \vec{c}_N, \quad \text{and} \quad \forall r \in \{N-1, \ldots, 0\} \colon \widehat{b}_r \stackrel{\text{def}}{=} 2\mathcal{M}(\widehat{b}_{r+1}) - \widehat{b}_{r+2} + \vec{c}_r \in \mathbb{R}^d \ ,$$

*and define the output as $\widehat{s}_N \stackrel{\text{def}}{=} \widehat{b}_0 - \mathcal{M}(\widehat{b}_1)$.*

If $\varepsilon = 0$, then $\widehat{s}_N = \vec{s}_N$. The following theorem gives an error analysis [6, Theorem 6.4].

**Theorem B.2** (stable Chebyshev sum). *For every $N \in \mathbb{N}^*$, suppose the eigenvalues of $\mathbf{M}$ are in $[a, b]$ and suppose there are parameters $C_U \geq 1, C_T \geq 1, \rho \geq 1, C_c \geq 0$ satisfying*

$$\forall k \in \{0, 1, \ldots, N\} \colon \left\{ \rho^k \|\vec{c}_k\| \leq C_c \bigwedge \forall x \in [a, b] \colon |\mathcal{T}_k(x)| \leq C_T \rho^k \text{ and } |\mathcal{U}_k(x)| \leq C_U \rho^k \right\} \ .$$

*Then, if the inexact backward recurrence in Def. B.1 is applied with $\varepsilon \leq \frac{1}{4NC_U}$, we have*

$$\|\widehat{s}_N - \vec{s}_N\| \leq \varepsilon \cdot 2(1 + 2NC_T)NC_U C_c \ .$$

Above, we have

**Claim B.3.** *Let $\mathcal{U}_t(x)$ be the $t$-th Chebyshev polynomial of the second kind, defined as:*

$$\mathcal{U}_0(x) \stackrel{\text{def}}{=} 1, \qquad \mathcal{U}_1(x) \stackrel{\text{def}}{=} 2x, \qquad \mathcal{U}_{t+1}(x) \stackrel{\text{def}}{=} 2x \cdot \mathcal{U}_t(x) - \mathcal{U}_{t-1}(x)$$

*then $\mathcal{U}_t(x)$ satisfies (see Trefethen [30]): $\frac{d}{dx}\mathcal{T}_t(x) = t\mathcal{U}_{t-1}(x)$ and thus*

$$\mathcal{U}_t(x) = \begin{cases} \in [-t, t] & \text{if } x \in [-1, 1]; \\ \frac{1}{2\sqrt{x^2-1}}\left[(x + \sqrt{x^2-1})^{t+1} - (x - \sqrt{x^2-1})^{t+1}\right] & \text{if } x > 1. \end{cases}$$

## B.2 Proof of Theorem 3

*Proof of Theorem 3.* We can without loss of generality assume $\delta \leq L$. For notation simplicity, let us denote

$$\mathbf{A} \stackrel{\text{def}}{=} \nabla^2 f(x_0), \ \mathbf{M} \stackrel{\text{def}}{=} \left(-\frac{1}{L}\nabla^2 f(x_0) + \left(1 - \frac{3\delta}{4L}\right)\mathbf{I}\right) \text{ and } \lambda \stackrel{\text{def}}{=} -\lambda_{\min}(\mathbf{A}).$$

Then, we know that the eigenvalues of $\mathbf{M}$ lie in $\left[-1, 1 + \frac{\lambda - 3\delta/4}{L}\right]$.

We wish to iteratively compute $x_{t+1} \approx x_0 + \mathcal{T}_t(\mathbf{M})\xi$, where $\mathcal{T}_t$ is the $t$-th Chebyshev polynomial of the first kind. However, we cannot multiply $\mathbf{M}$ to vectors (because we are not allowed to use Hessian-vector products). We define

$$\mathcal{M}(y) \stackrel{\text{def}}{=} -\frac{1}{L}\left(\nabla f(x_0 + y) - \nabla f(x_0)\right) + \left(1 - \frac{3\delta}{4L}\right)y \ .$$

and shall use it to approximate $\mathbf{M}y$ and then apply backward recurrence

$$y_0 = 0, \quad y_1 = \xi, \quad y_t = 2\mathcal{M}(y_{t-1}) - y_{t-2} \ .$$

If we set $x_{t+1} = x_0 + y_{t+1} - \mathcal{M}(y_t)$, following Def. B.1, it satisfies $x_{t+1} - x_0 \approx \mathcal{T}_t(\mathbf{M})\xi$.

Now, letting $x_{t+1}^* \stackrel{\text{def}}{=} x_0 + \mathcal{T}_t(\mathbf{M})\xi$ be the exact solution, we wish to bound the error $\|x_{t+1} - x_{t+1}^*\|_2$. Throughout the iterations of `Neon2det`, we have

$$y_t = 2\mathcal{M}(y_{t-1}) - y_{t-2} = 2(x_0 - x_t + y_t) - y_{t-2} \implies y_t - y_{t-2} = 2(x_t - x_0) \ .$$

Since we have $\|x_t - x_0\|_2 \leq r$ for each $t$ before termination, we know $\|y_t\|_2 \leq 2tr$. Using this upper bound we can approximate Hessian-vector product by gradient difference. Lemma A.1 gives us

$$\|\mathcal{M}(y_t) - \mathbf{M}y_t\|_2 \leq \frac{L_2}{L}\|y_t\|_2^2 \leq \frac{2L_2 rt}{L}\|y_t\|_2 \ .$$

Now, recall from Claim 4.1 that

$$\mathcal{T}_t(x) \in \begin{cases} [-1, 1] & \text{if } x \in [-1, 1]; \\ \left[\frac{1}{2}\left(x + \sqrt{x^2-1}\right)^t, \left(x + \sqrt{x^2-1}\right)^t\right] & \text{if } x > 1. \end{cases}$$

On the other hand, we have for every $x > 1$, $a = x + \sqrt{x^2-1}$, and $b = x - \sqrt{x^2-1}$, it satisfies

$$\mathcal{U}_t(x) = \frac{1}{a-b}\left(a^{t+1} - b^{t+1}\right) = \sum_{i=0}^t a^i b^{t-i} \leq (t+1)a^t \ .$$

Thus, we can apply Theorem B.2 with the eigenvalues of $\mathbf{M}$ in $[a, b] = \left[0, 1 + \frac{\lambda - 3\delta/4}{L}\right]$ and

$$\rho \stackrel{\text{def}}{=} \max\left\{1 + \frac{\lambda - 3\delta/4}{L} + \sqrt{2\frac{\lambda - 3\delta/4}{L} + \frac{(\lambda - 3\delta/4)^2}{L^2}}, 1\right\}, \quad C_c = \rho^t \sigma, \quad C_T = 2, C_U = t + 2 \ .$$

Theorem B.2 tells us that, for every $t$ before termination,

$$\|x_{t+1}^* - x_{t+1}\|_2 \leq \frac{40 L_2 r t^4 \rho^t \sigma}{L} \ .$$

In order to prove Theorem 3, in the rest of the proof, it suffices for us to show that, if $\lambda_{\min}(\nabla^2 f(x_0)) \leq -\delta$, then with probability at least $1 - p$, it satisfies $v \neq \bot$, $\|v\|_2 = 1$, and $v^\top \nabla^2 f(x_0) v \leq -\frac{1}{2}\delta$. In other words, we can assume $\lambda \geq \delta$.

The value $\lambda \geq \delta$ implies $\rho \geq 1 + \frac{1}{2}\sqrt{\frac{\delta}{L}} > 1$, so we can let

$$T_1 \stackrel{\text{def}}{=} \frac{\log \frac{4dr}{p\sigma}}{\log \rho} \leq T \ .$$

By Claim 4.1, we know that $\|\mathcal{T}_{T_1}(\mathbf{M})\|_2 \geq \frac{1}{2}\rho^{T_1} = \frac{2dr}{p\sigma}$. Thus, with probability at least $1 - p$, $\|x_{T_1+1}^* - x_0\|_2 = \|\mathcal{T}_{T_1}(\mathbf{M})\xi\|_2 \geq 2r$. Moreover, at iteration $T_1$, we have:

$$\|x_{T_1+1}^* - x_{T_1+1}\|_2 \leq \frac{40 L_2 r T_1^4 \rho^{T_1} \sigma}{L} \leq \frac{40 L_2 r T_1^4 \sigma}{L} \times \frac{4dr}{p\sigma} \leq \frac{256 d L_2 T_1^4}{L} \frac{r^2}{p} \stackrel{\textcircled{\tiny 1}}{\leq} \frac{\delta}{100L} r \leq \frac{1}{16} r \ .$$

Above, ① uses the fact that $r \leq \frac{\delta p}{25600 d L_2 T_1^4}$. This means $\|x_{T_1+1} - x_0\|_2 \geq r$ so the algorithm must terminate before iteration $T_1 \leq T$.

On the other hand, since $\|\mathcal{T}_t(\mathbf{M})\|_2 \leq \rho^t$, we know that the algorithm will not terminate until $t \geq T_0$ for

$$T_0 \stackrel{\text{def}}{=} \frac{\log \frac{r}{2\sigma}}{\log \rho} \ .$$

At the time $t \geq T_0$ of termination, define $\rho' = 1 + \frac{\lambda - 3\delta/4}{L}$, by the property of Chebyshev polynomial Claim 4.1, we know

1. $\mathcal{T}_t(\rho') \geq \frac{1}{2}\rho^t \geq \frac{1}{2}\rho^{T_0} \geq \frac{r}{4\sigma} = (d/p)^{\Theta(1)}$.
2. $\forall x \in [-1,1], \mathcal{T}_t(x) \in [-1,1]$.

Since all the eigenvalues of $\mathbf{A}$ that are $\geq -3/4\delta$ are mapped to the eigenvalues of $\mathbf{M}$ that are in $[-1,1]$, and the smallest eigenvalue of $\mathbf{A}$ is mapped to the eigenvalue $\rho'$ of $\mathbf{M}$. So we have, with probability at least $1-p$, letting $v_t \stackrel{\text{def}}{=} x_{t+1}^* - x_0$ it satisfies

$$\frac{v_t^\top \mathbf{A} v_t}{\|v_t\|_2^2} \leq -\frac{5}{8}\delta \ .$$

Therefore, denoting by $z_t \stackrel{\text{def}}{=} x_{t+1} - x_0$, we have

$$\frac{z_t^\top \mathbf{A} z_t}{\|z_t\|_2^2} = \frac{\|v_t\|_2^2}{\|z_t\|_2^2} \cdot \frac{z_t^\top \mathbf{A} z_t}{\|v_t\|_2^2} \leq \frac{\|v_t\|_2^2}{\|z_t\|_2^2} \frac{v_t^\top \mathbf{A} v_t + 4L\|z_t - v_t\|_2 \|v_t\|_2}{\|v_t\|_2^2}$$
$$\leq \frac{\|v_t\|_2^2}{\|z_t\|_2^2} \left( \frac{v_t^\top \mathbf{A} v_t}{\|v_t\|_2^2} + \frac{4L\|v_t - z_t\|_2}{\|v_t\|_2} \right) \leq \frac{\|v_t\|_2^2}{\|z_t\|_2^2} \left( -\frac{5}{8}\delta + \frac{1}{25}\delta \right)$$
$$\leq \frac{15}{16} \left( -\frac{5}{8}\delta + \frac{1}{25}\delta \right) \leq -\frac{51}{100}\delta \ .$$

This finishes the proof because we have shown that, with probability at least $1-p$, the output $v = \frac{z_t}{\|z_t\|}$ satisfies and $v^\top \nabla^2 f(x_0) v \leq -\frac{1}{2}\delta$. □

## C   Proofs for Section 6: Applications of Neon2

### C.1   Auxiliary Claims

**Claim C.1.** *For any $x$, using $O((\frac{\mathcal{V}}{\varepsilon^2} + 1)\log\frac{1}{p})$ stochastic gradients, we can decide*

$$\text{with probability } 1-p: \quad \text{either } \|\nabla f(x)\| \geq \frac{\varepsilon}{2} \quad \text{or} \quad \|\nabla f(x)\| \leq \varepsilon \ .$$

*Proof.* Suppose we generate $m = O(\log\frac{1}{p})$ random uniform subsets $S_1, \ldots, S_m$ of $[n]$, each of cardinality $B = \max\{\frac{32\varepsilon^2}{\mathcal{V}}, 1\}$. Then, denoting by $v_j = \frac{1}{B}\sum_{i \in S_j} \nabla f_i(x)$, we have according to Fact 2.2 that $\mathbb{E}_{S_j}\left[\|v_j - \nabla f(x)\|^2\right] \leq \frac{\mathcal{V}}{B} = \frac{\varepsilon^2}{32}$. In other words, with probability at least $1/2$ over the randomness of $S_j$, we have $\big|\|v_j\| - \|\nabla f(x)\|\big| \leq \|v_j - \nabla f(x)\| \leq \frac{\varepsilon}{4}$. Since $m = O(\log\frac{1}{p})$, we have with probability at least $1-p$, it satisfies that at least $m/2 + 1$ of the vectors $v_j$ satisfy $\big|\|v_j\| - \|\nabla f(x)\|\big| \leq \frac{\varepsilon}{4}$. Now, if we select $v^* = v_j$ where $j \in [m]$ is the index that gives the median value of $\|v_j\|$, then it satisfies $\big|\|v_j\| - \|\nabla f(x)\|\big| \leq \frac{\varepsilon}{4}$. Finally, we can check if $\|v_j\| \leq \frac{3\varepsilon}{4}$. If so, then we conclude that $\|\nabla f(x)\| \leq \varepsilon$, and if not, we conclude that $\|\nabla f(x)\| \geq \frac{\varepsilon}{2}$. □

**Claim C.2.** *If $v$ is a unit vector and $v^\top \nabla^2 f(y) v \leq -\frac{\delta}{2}$, suppose we choose $y' = y \pm \frac{\delta}{L_2}v$ where the sign is random, then $f(y) - \mathbb{E}[f(y')] \geq \frac{\delta^3}{12L_2^2}$.*

*Proof.* Letting $\eta = \frac{\delta}{L_2}$, then by the second-order smoothness,

$$f(y) - \mathbb{E}[f(y')] \geq \mathbb{E}\left[\langle \nabla f(y), y - y'\rangle - \frac{1}{2}(y-y')^\top \nabla^2 f(y)(y-y') - \frac{L_2}{6}\|y - y'\|^3\right]$$
$$= -\frac{\eta^2}{2} v^\top \nabla^2 f(y) v - \frac{L_2 \eta^3}{6}\|v\|^3 \geq \frac{\eta^2 \delta}{4} - \frac{L_2 \eta^3}{6} = \frac{\delta^3}{12L_2^2} \ .$$
□

## C.2 Proof of Theorem 7a

*Proof of Theorem 7a.* Since both estimating $\|\nabla f(x_t)\|$ in Line 5 (see Claim C.1) and invoking Neon2$^{\text{online}}$ (see Theorem 1) succeed with high probability, we can assume that they always succeed. This means whenever we output $x_t$ in an iteration, it already satisfies $\|\nabla f(x_t)\| \leq \varepsilon$ and $\nabla^2 f(x_t) \succeq -\delta I$. Therefore, it remains to show that the algorithm must output some $x_t$ in an iteration, as well as to compute the final complexity.

Recall (from classical SGD theory) if we update $x_{t+1/2} \leftarrow x_t - \frac{\alpha}{|S|} \sum_{i \in S} \nabla f_i(x_t)$ where $\alpha > 0$ is the learning rate and $|S| = B$, then

$$f(x_t) - \mathbb{E}_S[f(x_{t+1/2})] \overset{\text{①}}{\geq} \mathbb{E}_S\left[\langle \nabla f(x_t), x_t - x_{t+1/2}\rangle - \frac{L}{2}\|x_t - x_{t+1/2}\|^2\right]$$

$$= \alpha \|\nabla f(x_t)\|^2 - \frac{\alpha^2 L}{2} \mathbb{E}_S\left[\Big\|\frac{1}{|S|}\sum_{i \in S}\nabla f_i(x_t)\Big\|^2\right]$$

$$= (\alpha - \frac{\alpha^2 L}{2})\|\nabla f(x_t)\|^2 - \frac{\alpha^2 L}{2}\mathbb{E}_S\left[\Big\|\nabla f(x_t) - \frac{1}{|S|}\sum_{i\in S}\nabla f_i(x_t)\Big\|^2\right]$$

$$\overset{\text{②}}{\geq} (\alpha - \frac{\alpha^2 L}{2})\|\nabla f(x_t)\|^2 - \frac{\alpha^2 L}{2}\frac{\mathcal{V}}{B} \ .$$

Above, ① is due to the smoothness of $f(\cdot)$ and ② is due to Fact 2.2. Now, if we choose $\alpha = \frac{1}{L}$ and $B = \max\{\frac{8\mathcal{V}}{\varepsilon^2}, 1\}$, then we have

$$f(x_t) - \mathbb{E}_S[f(x_{t+1/2})] \geq \frac{\alpha}{2}\left(\|\nabla f(x_t)\|^2 - \frac{\varepsilon^2}{8}\right) \ . \tag{C.1}$$

In other words, as long as Line 6 is reached, we have $f(x_t) - \mathbb{E}[f(x_{t+1})] \geq \Omega(\varepsilon^2/L)$. On the other hand, whenever Line 10 is reached, then we must have $v^\top \nabla^2 f(y_0) v \leq -\frac{\delta}{2}$. By Claim C.2, we must have $f(x_t) - \mathbb{E}[f(x_{t+1})] \geq \Omega(\delta^3/L_2^2)$.

In sum, if we choose $K = O\big(\frac{L_2^2 \Delta_f}{\delta^3} + \frac{L\Delta_f}{\varepsilon^2}\big)$, then the algorithm must terminate and return $x_t$ in one of its iterations. This ensures that Line 13 will not be reached. As for the total complexity, we note that each iteration of Neon2+SGD is dominated by $\widetilde{O}(B) = \widetilde{O}(\frac{\mathcal{V}}{\varepsilon^2} + 1)$ stochastic gradient computations in Line 4 and Line 5, totaling $\widetilde{O}((\frac{\mathcal{V}}{\varepsilon^2} + 1)K)$, as well as $\widetilde{O}(\frac{L^2}{\delta^2})$ stochastic gradient computations by Neon2$^{\text{online}}$, but the latter will not happen for more than $O(\frac{L_2^2 \Delta_f}{\delta^3})$ times. Therefore, the total gradient complexity is

$$\widetilde{O}\Big((\frac{\mathcal{V}}{\varepsilon^2}+1)K + \frac{L^2}{\delta^2}\frac{L_2^2\Delta_f}{\delta^3}\Big) = \widetilde{O}\Big((\frac{\mathcal{V}}{\varepsilon^2}+1)\big(\frac{L_2^2\Delta_f}{\delta^3}+\frac{L\Delta_f}{\varepsilon^2}\big) + \frac{L^2}{\delta^2}\frac{L_2^2\Delta_f}{\delta^3}\Big) \ . \quad \square$$

## C.3 Proof of Theorem 7b

*Proof of Theorem 7b.* We first note in the special case $b = \Theta(\frac{(\varepsilon^2+\mathcal{V})\varepsilon^4 L_2^6}{\delta^9 L^3}) \geq B$, or equivalently $\delta^3 \leq O(\frac{L_2^2 \varepsilon^2}{L})$, Theorem 7a gives us gradient complexity $T = \widetilde{O}\Big(\frac{\mathcal{V}}{\varepsilon^2}\frac{L_2^2\Delta_f}{\delta^3} + \frac{L^2}{\delta^2}\frac{L_2^2\Delta_f}{\delta^3}\Big)$ so we are done.

Therefore, in the rest of the proof we assume $\Theta(\frac{(\varepsilon^2+\mathcal{V})\varepsilon^4 L_2^6}{\delta^9 L^3}) < B$ and thus $b \leq B$ is well defined. Since both estimating $\|\nabla f(x_t)\|$ in Line 6 (see Claim C.1) and invoking Neon2$^{\text{online}}$ (see Theorem 1) succeed with high probability, we can assume that they always succeed. This means whenever we output $x_{t+1/2}$ in an iteration, it already satisfies $\|\nabla f(x_{t+1/2})\| \leq \varepsilon$ and $\nabla^2 f(x_{t+1/2}) \succeq -\delta \mathbf{I}$. Therefore, it remains to show that the algorithm must output some $x_t$ in an iteration, as well as to compute the final complexity.

For analysis purpose, let us assume that, whenever the algorithm reaches $v = \bot$ (in Line 10), it does not immediately halt and instead sets $x_{t+1} = x_{t+1/2}$. This modification ensures that random variables $x_t$ and $x_{t+1/2}$ are well defined for $t = 0, 1, \ldots, K-1$.

Let $N_1$ and $N_2$ respectively be the number of times we reach Line 9 (so we invoke `Neon2`[online]) and the number of times we reach Line 11 (so we update $x_{t+1} = x_{t+1/2} \pm \frac{\delta}{L_2} v$). Both $N_1$ and $N_2$ are random variables and it satisfies $N_1 \geq N_2$. To prove that `Neon2+SCSG` outputs in an iteration, we need to prove $N_1 > N_2$ holds at least with constant probability.

Let us apply the key SCSG lemma (Lemma 6.2) for an epoch with size $B = \max\{1, \frac{48\mathcal{V}}{\varepsilon^2}\}$ and mini-batch size $b \geq 1$, we have

$$\mathbb{E}\big[\|\nabla f(x_{t+1/2})\|^2\big] \leq C \cdot L(b/B)^{1/3} \big(f(x_t) - \mathbb{E}[f(x_{t+1/2})]\big) + \frac{\varepsilon^2}{8} . \tag{C.2}$$

Now,

- if $\|\nabla f(x_{t+1/2})\| \geq \frac{\varepsilon}{2}$ (so Line 7 is reached), we have $x_{t+1} = x_{t+1/2}$;
- if $v = \bot$ holds (so Line 9 is reached), we have virtually set $x_{t+1} = x_{t+1/2}$ for analysis purpose;
- if $v \neq \bot$ (so Line 11 is reached), we have $\frac{\delta^3}{12 L_2^2} \leq f(x_{t+1/2}) - \mathbb{E}[f(x_{t+1})]$ .

Note that the third case $v \neq \bot$ happens for $N_2$ times. Therefore, combining the three cases together with (C.2), we have

$$\frac{B^{1/3}}{CLb^{1/3}} \mathbb{E}\Big[\sum_{t=0}^{K-1} \big(\|\nabla f(x_{t+1/2})\|^2 - \frac{\varepsilon^2}{8}\}\big)\Big] + \frac{\delta^3}{12 L_2^2} \cdot \mathbb{E}[N_2] \leq \Delta_f .$$

On one hand, since we have chosen $K$ such that $K \geq \Omega\big(\frac{Lb^{1/3}\Delta_f}{\varepsilon^2 B^{1/3}}\big) = \Omega\big(\frac{Lb^{1/3}\Delta_f}{\varepsilon^2(1+\mathcal{V}/\varepsilon^2)^{1/3}}\big)$, by Markov bound (ignoring $\mathbb{E}[N_2]$), with probability at least $5/6$, it satisfies $\sum_{t=0}^{K-1} \|\nabla f(x_{t+1/2})\|^2 \leq \frac{\varepsilon^2}{4} K$. As a consequence, at least half of the indices $t = 0, 1, \ldots, K-1$ will satisfy $\|\nabla f(x_{t+1/2})\| \leq \frac{\varepsilon}{2}$. This means we have $N_1 \geq K/2$.

On the other hand, we have $\frac{\delta^3}{12 L_2^2} \cdot \mathbb{E}[N_2] \leq \Delta_f + \frac{KB^{1/3}\varepsilon^2}{8CLb^{1/3}}$. Since $K \geq \Omega\big(\frac{Lb^{1/3}\Delta_f}{\varepsilon^2 B^{1/3}}\big) = \Omega\big(\frac{Lb^{1/3}\Delta_f}{\varepsilon^2(1+\mathcal{V}/\varepsilon^2)^{1/3}}\big)$, we have $\frac{\delta^3}{12 L_2^2} \cdot \mathbb{E}[N_2] \leq \frac{KB^{1/3}\varepsilon^2}{4CLb^{1/3}}$. As long as $B \leq O\big(\frac{\delta^9 L^3 b}{\varepsilon^6 L_2^6}\big)$ or equivalently $b \geq \Omega\big(\frac{(\varepsilon^2+\mathcal{V})\varepsilon^4 L_2^6}{\delta^9 L^3}\big)$, we have $\mathbb{E}[N_2] < K/12$. Therefore, with probability at least $5/6$, it satisfies $N_2 < K/2$.

Since $N_1 > N_2$, this means the algorithm must terminate and output some $x_{t+1/2}$ in an iteration, with probability at least $2/3$.

Finally, the per-iteration complexity of `Neon2+SCSG` is dominated by $\widetilde{O}(B)$ stochastic gradient computations per iteration for both SCSG and estimating $\|\nabla f(x_{t+1/2})\|$, as well as $\widetilde{O}(L^2/\delta^2)$ for invoking `Neon2`[online]. This totals to gradient complexity

$$\widetilde{O}\Big(K\big(B + \frac{L^2}{\delta^2}\big)\Big) = \widetilde{O}\Big(\big(\frac{Lb^{1/3}\Delta_f}{\varepsilon^2(1+\mathcal{V}/\varepsilon^2)^{1/3}}\big)\big(\frac{\mathcal{V}}{\varepsilon^2} + \frac{L^2}{\delta^2}\big)\Big)$$

$$= \widetilde{O}\Big(\big(\frac{Lb^{1/3}\Delta_f}{\varepsilon^{4/3}\mathcal{V}^{1/3}}\big)\big(\frac{\mathcal{V}}{\varepsilon^2} + \frac{L^2}{\delta^2}\big) + \frac{L\Delta_f}{\varepsilon^2}\frac{L^2}{\delta^2}\Big)$$

$$= \widetilde{O}\Big(\big(\frac{Lb^{1/3}\Delta_f}{\varepsilon^{4/3}\mathcal{V}^{1/3}}\big)\big(\frac{\mathcal{V}}{\varepsilon^2} + \frac{L^2}{\delta^2}\big) + \frac{L\Delta_f}{\varepsilon^2}\frac{L^2}{\delta^2}\Big)$$

$$= \widetilde{O}\Big(\big(\frac{L\Delta_f}{\varepsilon^{4/3}\mathcal{V}^{1/3}} + \frac{L_2^2\Delta_f}{\delta^3}\big)\big(\frac{\mathcal{V}}{\varepsilon^2} + \frac{L^2}{\delta^2}\big) + \frac{L\Delta_f}{\varepsilon^2}\frac{L^2}{\delta^2}\Big) . \qquad \square$$

## References

[1] Naman Agarwal, Zeyuan Allen-Zhu, Brian Bullins, Elad Hazan, and Tengyu Ma. Finding Approximate Local Minima for Nonconvex Optimization in Linear Time. In *STOC*, 2017. Full version available at

...
http://arxiv.org/abs/1611.01146.

[2] Zeyuan Allen-Zhu. Natasha 2: Faster Non-Convex Optimization Than SGD. *ArXiv e-prints*, abs/1708.08694, August 2017. Full version available at http://arxiv.org/abs/1708.08694.

[3] Zeyuan Allen-Zhu. Katyusha X: Practical Momentum Method for Stochastic Sum-of-Nonconvex Optimization. *ArXiv e-prints*, abs/1802.03866, February 2018. Full version available at http://arxiv.org/abs/1802.03866.

[4] Zeyuan Allen-Zhu and Elad Hazan. Variance Reduction for Faster Non-Convex Optimization. In *ICML*, 2016. Full version available at http://arxiv.org/abs/1603.05643.

[5] Zeyuan Allen-Zhu and Yuanzhi Li. LazySVD: Even Faster SVD Decomposition Yet Without Agonizing Pain. In *NIPS*, 2016. Full version available at http://arxiv.org/abs/1607.03463.

[6] Zeyuan Allen-Zhu and Yuanzhi Li. Faster Principal Component Regression and Stable Matrix Chebyshev Approximation. In *Proceedings of the 34th International Conference on Machine Learning*, ICML '17, 2017.

[7] Zeyuan Allen-Zhu and Yuanzhi Li. Follow the Compressed Leader: Faster Online Learning of Eigenvectors and Faster MMWU. In *ICML*, 2017. Full version available at http://arxiv.org/abs/1701.01722.

[8] Yair Carmon, John C. Duchi, Oliver Hinder, and Aaron Sidford. Accelerated Methods for Non-Convex Optimization. *ArXiv e-prints*, abs/1611.00756, November 2016.

[9] Yair Carmon, Oliver Hinder, John C. Duchi, and Aaron Sidford. "Convex Until Proven Guilty": Dimension-Free Acceleration of Gradient Descent on Non-Convex Functions. In *ICML*, 2017.

[10] Anna Choromanska, Mikael Henaff, Michael Mathieu, Gérard Ben Arous, and Yann LeCun. The loss surfaces of multilayer networks. In *AISTATS*, 2015.

[11] Yann N Dauphin, Razvan Pascanu, Caglar Gulcehre, Kyunghyun Cho, Surya Ganguli, and Yoshua Bengio. Identifying and attacking the saddle point problem in high-dimensional non-convex optimization. In *NIPS*, pages 2933–2941, 2014.

[12] Aaron Defazio, Francis Bach, and Simon Lacoste-Julien. SAGA: A Fast Incremental Gradient Method With Support for Non-Strongly Convex Composite Objectives. In *NIPS*, 2014.

[13] Dan Garber, Elad Hazan, Chi Jin, Sham M. Kakade, Cameron Musco, Praneeth Netrapalli, and Aaron Sidford. Robust shift-and-invert preconditioning: Faster and more sample efficient algorithms for eigenvector computation. In *ICML*, 2016.

[14] Rong Ge, Furong Huang, Chi Jin, and Yang Yuan. Escaping from saddle points—online stochastic gradient for tensor decomposition. In *Proceedings of the 28th Annual Conference on Learning Theory*, COLT 2015, 2015.

[15] I. J. Goodfellow, O. Vinyals, and A. M. Saxe. Qualitatively characterizing neural network optimization problems. *ArXiv e-prints*, December 2014.

[16] Chi Jin, Rong Ge, Praneeth Netrapalli, Sham M Kakade, and Michael I Jordan. How to Escape Saddle Points Efficiently. In *ICML*, 2017.

[17] Rie Johnson and Tong Zhang. Accelerating stochastic gradient descent using predictive variance reduction. In *Advances in Neural Information Processing Systems*, NIPS 2013, pages 315–323, 2013.

[18] Cornelius Lanczos. An iteration method for the solution of the eigenvalue problem of linear differential and integral operators. *Journal of Research of the National Bureau of Standards*, 45(4), 1950.

[19] Lihua Lei, Cheng Ju, Jianbo Chen, and Michael I Jordan. Nonconvex Finite-Sum Optimization Via SCSG Methods. In *NIPS*, 2017.

[20] Yurii Nesterov. *Introductory Lectures on Convex Programming Volume: A Basic course*, volume I. Kluwer Academic Publishers, 2004. ISBN 1402075537.

[21] Yurii Nesterov and Boris T. Polyak. Cubic regularization of newton method and its global performance. *Mathematical Programming*, 108(1):177–205, 2006.

[22] Erkki Oja. Simplified neuron model as a principal component analyzer. *Journal of mathematical biology*, 15(3):267–273, 1982.

[23] Barak A Pearlmutter. Fast exact multiplication by the hessian. *Neural computation*, 6(1):147–160, 1994.

[24] Sashank J. Reddi, Ahmed Hefny, Suvrit Sra, Barnabas Poczos, and Alex Smola. Stochastic variance reduction for nonconvex optimization. In *ICML*, 2016.

[25] Sashank J Reddi, Manzil Zaheer, Suvrit Sra, Barnabas Poczos, Francis Bach, Ruslan Salakhutdinov, and Alexander J Smola. A generic approach for escaping saddle points. *ArXiv e-prints*, abs/1709.01434, September 2017.

[26] Youcef Saad. *Numerical methods for large eigenvalue problems*. Manchester University Press, 1992.

[27] Mark Schmidt, Nicolas Le Roux, and Francis Bach. Minimizing finite sums with the stochastic average gradient. *ArXiv e-prints*, abs/1309.2388, September 2013. Preliminary version appeared in NIPS 2012.

[28] Nicol N Schraudolph. Fast curvature matrix-vector products for second-order gradient descent. *Neural computation*, 14(7):1723–1738, 2002.

[29] Shai Shalev-Shwartz. SDCA without Duality, Regularization, and Individual Convexity. In *ICML*, 2016.

[30] Lloyd N. Trefethen. *Approximation Theory and Approximation Practice*. SIAM, 2013.

[31] Yi Xu and Tianbao Yang. First-order Stochastic Algorithms for Escaping From Saddle Points in Almost Linear Time. *ArXiv e-prints*, abs/1711.01944, November 2017.